\documentclass[runningheads]{llncs}
\usepackage[T1]{fontenc}
\usepackage{graphicx}
\usepackage{booktabs}
\usepackage[misc]{ifsym}

\usepackage{mwe}

\usepackage{graphicx}%
\usepackage{multirow}%
\usepackage{amsmath,amssymb,amsfonts}%
\usepackage{mathrsfs}%
\usepackage[title]{appendix}%
\usepackage{xcolor}%
\usepackage{textcomp}%
\usepackage{manyfoot}%
\usepackage{booktabs}%
\usepackage{algorithm}%
\usepackage{algorithmicx}%
\usepackage{algpseudocode}%
\usepackage{listings}%

\usepackage{array}
\usepackage{bm}
\usepackage{subcaption}
\usepackage{hyperref}

\usepackage{pgf}
\usepackage{tikz}
\usetikzlibrary{shapes}

\newcolumntype{H}{>{\setbox0=\hbox\bgroup}c<{\egroup}@{}}

\DeclareMathOperator*{\argmin}{argmin}

\DeclareMathOperator*{\ch}{ch}

\DeclareMathOperator*{\scope}{sc}

\newcommand{\X}{\mathbf{X}}
\newcommand{\x}{\mathbf{x}}

\newcommand{\xxs}{\bm{\mathcal{X}}}

\newcommand{\graph}{\mathcal{G}}

\newcommand{\node}{v}
\newcommand{\nodec}{u}

\newcommand{\w}{w}

\newcommand{\noto}{{\ensuremath{\lnot o}}}

\begin{document}

\title{Probabilistic Circuits with Constraints via Convex Optimization}


\author{Soroush Ghandi\inst{1} \and Benjamin Quost\inst{2} \and Cassio de Campos\inst{1}}

\authorrunning{Ghandi et al.}

\institute{Eindhoven Unisversity of Technology \and University of Technology of Compiègne}

\maketitle              

\begin{abstract}
This work addresses integrating probabilistic propositional logic constraints into the distribution encoded by a probabilistic circuit (PC). PCs are a class of tractable models that allow efficient computations (such as conditional and marginal probabilities) while achieving state-of-the-art performance in some domains. The proposed approach takes both a PC and constraints as inputs, and outputs a new PC that satisfies the constraints. This is done efficiently via convex optimization without the need to retrain the entire model. Empirical evaluations indicate that the combination of constraints and PCs can have multiple use cases, including the improvement of model performance under scarce or incomplete data, as well as the enforcement of machine learning fairness measures into the model without compromising model fitness. We believe that these ideas will open possibilities for multiple other applications involving the combination of logics and deep probabilistic models.

\end{abstract}

\section{Introduction}\label{sec1}

Generative probabilistic models typically aim to learn the joint probability distribution of data, in order to perform probabilistic inference and answer queries of interest. However, not all the probabilistic models are the same in that regard. Models like variational autoencoders (VAEs) \cite{kingma2013auto} and generative adversarial networks (GANs) \cite{goodfellow2020generative} possess exceptional modeling prowess; nevertheless, their ability to perform probabilistic inference such as marginalization and conditioning remains rather limited, due to tractability issues. 

In contrast, \emph{tractable} probabilistic models, such as probabilistic circuits (PCs), including the prominent sum-product networks (SPNs)  \cite{poon2011sum,sanchez2021sum}, allow for a wider range of exact inferences, arguably at the expense of some fitting power. PCs fall within the family of probabilistic graphical models (PGMs), a class of models using a graph-based representation to encode high-dimensional distributions \cite{koller2009probabilistic}. Unlike Bayesian networks, which have a notoriously high complexity for general queries \cite{decampos2011new}, 
PCs can produce several types of inferences in polynomial time under arguably mild assumptions \cite{vergari2021compositional}.

Learning a PC from data $\mathcal{D}$ is defined as specifying a PC that represents the probability distribution underlying $\mathcal{D}$.
This active line of research has seen several meaningful proposals in the past few years, such as \cite{Adel2015,dennis2012learning,di2017fast,gens2013learning,hsu2017online,kalra2018online,lee2014non} and \cite{liang2017learning,liu2021tractable,molina2018mixed,peharz2013greedy,peharz2020einsum,rahman2016merging,rahman2019look,rooshenas2014,trapp2019bayesian,vergari2015simplifying,vergari2019automatic}, but remains nevertheless open, given the difficulty of the task which involves both structure and parameter learning.

We address here the issue of \emph{enhancing} a PC learned from data by using additional information and/or learning goals. To this end, we propose an approach for combining the PC with probabilistic propositional logic (PPL) constraints. 
More specifically, the approach takes a learned PC and updates (some of) its parameters in order to enforce the PPL constraints globally in the represented distribution. Our strategy can be seen as a ``post-learning'' method, which gives the advantages of versatility (existing models need not be retrained) as well as modularity: one may train a PC using any algorithm, as long as the resulting network 
keeps dependent variables (which may appear together in the same PPL constraint) together within the model; that is, they cannot appear factorized in the graph (further details are given in Section \ref{sec:CPC}). 
This allows to build convex optimization problems (more precisely, constrained KL-divergence solvers) over parts of the distribution encoded in the PC so as to improve the corresponding model parameters via an efficient tractable method.

The benefits of having user-specified constraints are multi-fold. In order to illustrate them, we employ PPL constraints in a few (non-exhaustive) scenarios: (1) we improve the quality of models by enforcing that the yielded model matches the empirical marginal distributions under situations of (a) scarce data or (b) missing data; (2) we enforce fairness constraints into the model while at the same time avoiding a decrease in fitness. Overall, the experiments indicate that using PPL constraints often yields a better model (without compromising efficiency or accuracy), which is likely possible because of typical over-parametrizations that current large machine learning models impose. We emphasize that these applications of constraints are only a few examples of possible use, as we believe there are many other possibilities ahead to be tried.

\section{Probabilistic circuits}\label{sec:PCs}

Probabilistic circuits (PCs) are a family of distribution representations facilitating many exact and efficient inference routines (see \cite{VanDenBroeck2019} for a nice introduction). A PC encodes a probabilistic model over a collection of variables $\X$; it is structured as a rooted directed acyclic graph $\graph$, containing three types of nodes: (i) distribution nodes, (ii) sum nodes, and (iii) product nodes.
Distribution nodes are the leaves of the graph $\graph$, while sum and product nodes are the internal nodes. 
Each distribution node (leaf) $\node$ computes a probability distribution over some subset $\X' \subseteq \X$, i.e. an integrable function $p_\node(\x') \colon \xxs' \to \mathbb{R}^{+}$ from the sample space of $\X'$ to the non-negative real numbers. 
The \emph{scope} of $v$ is the set of variables $\scope(\node) := \X'$ over which the leaf computes a distribution. 
The scope of any internal node $\node$ (sum or product) is recursively defined as $\scope(\node) = \cup_{\nodec \in \ch(\node)} \scope(\nodec)$, where $\ch(\node)$ is the set containing the children of $\node$. 
Sum nodes compute convex combinations over their children, i.e.~if $\node$ is a sum node, then $\node$ computes $\node(\x) = \sum_{\nodec \in \ch(\node)} \w_{\node,\nodec} \nodec(\x)$, where $\w_{\node,\nodec} \geq 0$. In a normalized PC, we have $\sum_{\nodec \in \ch(\node)} \w_{\node,\nodec} = 1$.
Product nodes compute the product over their children, i.e.~if $\node$ is a product node, then $\node(\x) = \prod_{\nodec \in \ch(\node)} \nodec(\x)$. 
The support of a node is the region where its associated function is strictly positive. 

The main feature of PCs is that they facilitate a wide range of \emph{tractable} inference routines, which go hand in hand with certain structural properties \cite{VanDenBroeck2019,Darwiche2003}: 
(i) a sum node $\node$ is called \emph{smooth} if its children have all the same scope: $\scope(\nodec) = \scope(\nodec')$, for any $\nodec, \nodec' \in \ch(\node)$; 
(ii) a product node $\node$ is called \emph{decomposable} if its children have non-overlapping scopes: $\scope(\nodec) \cap \scope(\nodec') = \emptyset$, for any $\nodec, \nodec' \in \ch(\node)$, $\nodec \not= \nodec'$; 
(iii) a node is \emph{consistent} if its support is non-empty.
A PC is smooth (respectively decomposable) if all its sum (respectively product) nodes are smooth (respectively decomposable). A PC is consistent if all its nodes are consistent. 
The distribution $p(\scope(\node))$ represented by a node $\node$ in the PC is the function computed by the rules of the previous paragraph, and can be evaluated with a feed-forward pass. 
In order to ensure the tractability of queries, we can rely on smoothness and decomposability, but we also need leaf distribution nodes to compute inferences efficiently. This is a reason for many proposed PCs in the literature to assume that leaf nodes are univariate with some known distribution, such as Bernoulli, categorical, Gaussian, etc. 
Now, assume that we wish to compute a \emph{marginal query}, that is, to evaluate the probability value over $\X_o \subset \X$ for evidence $\X_o = \x_o$, while marginalising $\X_\noto = \X \setminus \X_o$. In smooth and decomposable PCs, this task reduces to performing marginalization at the leaves \cite{Peharz2015}: for each leaf $\node$, one marginalizes $\scope(\node) \cap \X_\noto$, and evaluates it for the values corresponding to $\scope(\node) \cap \X_o$. 
The desired marginal $p_{\X_o}(\x_o)$ results from evaluating internal nodes as in computing the complete distribution. 
Smoothness and consistency are sufficient to guarantee that the function of a PC represents a distribution. We also assume PCs are normalized. 

\definecolor{blu}{RGB}{199, 206, 234}
\definecolor{gre}{RGB}{181, 234, 215}
\definecolor{re}{RGB}{255, 154, 162}
\definecolor{ore}{RGB}{255, 218, 193}
\definecolor{lgr}{RGB}{226, 240, 203}
\definecolor{mel}{RGB}{255, 183, 178}

\begin{figure}[ht!]
\centering
\begin{tikzpicture}[scale=0.8,every node/.style={transform shape},
    roundnode/.style={minimum size=7.5mm,
                      inner sep=0}, 
    cross/.style={minimum size=5mm, fill=lgr,
        path picture={
            \draw[black] (path picture bounding box.south east) -- (path picture bounding box.north west) (path picture bounding box.south west) -- (path picture bounding box.north east);
        }
    },
    sum/.style={minimum size=5mm, fill=blu,
        path picture={
            \draw[black] (path picture bounding box.south) -- (path picture bounding box.north) (path picture bounding box.west) -- (path picture bounding box.east);
        }
    },
    gauss/.style={minimum size=5mm, fill=ore,
        path picture={
            \draw[black] plot[domain=-.15:.15] ({\x},{exp(-200*\x*\x -2.)});
        }
    },
    dt/.style={minimum size=5mm, fill=white
    },
    line/.style={
      draw,thick,
      -latex',
      shorten >=2pt
    },
    cloud/.style={
      draw=red,
      thick,
      ellipse,
      fill=red!20,
      minimum height=1em
    }
]

    \node[draw, circle, sum] (S0) at (7, 3-3.75) { };

    \node[draw, circle, cross] (P11) at (5, 2-3.75) { };
    \node[draw, circle, cross] (P12) at (9, 2-3.75) { };
    \node[draw, circle, gauss] (G11) at (4, 2-3.75) { };
    \node (b) at (3.1, 2-3.75) {\footnotesize $p_a(X_1)$};

    \node[draw, circle, sum] (S21) at (5, 1-3.75) { };
    \node[draw, circle, sum] (S22) at (8, 1-3.75) { };
    \node[draw, circle, gauss] (G21) at (10, 1-3.75) { };
    \node (b) at (10.9, 1-3.75) {\footnotesize $p_i(X_2)$};

    \node[draw, circle, cross] (P31) at (4, 0-3.75) { };
    \node[draw, circle, cross] (P32) at (6, 0-3.75) { };
    \node[draw, circle, cross] (P33) at (8, 0-3.75) { };
    \node[draw, circle, cross] (P34) at (10, 0-3.75) { };
    
    \node[draw, circle, gauss] (G40) at (3, -1-3.75) { };
    \node[draw, circle, gauss] (G41) at (4, -1-3.75) {};
    \node[draw, circle, gauss] (G42) at (6, -1-3.75) { };
    \node[draw, circle, gauss] (G43) at (7, -1-3.75) { };
    \node[draw, circle, gauss] (G44) at (8, -1-3.75) { };
    \node[draw, circle, gauss] (G45) at (10, -1-3.75) { };
    \node[draw, circle, gauss] (G46) at (11, -1-3.75) { };
    
    \node (b) at (2.9, -5.3) {\footnotesize $p_b(X_2)$};
    \node (b) at (5.9, -1.4-3.9) {\footnotesize $p_d(X_2)$};
    \node (b) at (4.1, -1.4-3.9) {\footnotesize $p_c(X_3)$};
    \node (b) at (7, -1.4-3.9) {\footnotesize $p_e(X_3)$};
    \node (b) at (8.1, -1.4-3.9) {\footnotesize $p_f(X_1)$};
    \node (b) at (9.9, -1.4-3.9) {\footnotesize $p_g(X_1)$};
    \node (b) at (11.1, -1.4-3.9) {\footnotesize $p_h(X_3)$};

    \draw[line width=0.3mm, ->, above] (S0.225) to node[black]{$.8$} (P11.45);
    \draw[line width=0.3mm, ->, above] (S0.315) to node[black]{$.2$} (P12.135);
    
    \draw[line width=0.3mm, ->, above] (P11.180) to node[black]{} (G11.0);
    \draw[line width=0.3mm, ->] (P11.270) -- (S21.90);
    \draw[line width=0.3mm, ->] (P12.225) -- (S22.45);
    \draw[line width=0.3mm, ->] (P12.315) -- (G21.135);
    
    \draw[line width=0.3mm, ->, above] (S21.225) to node[black]{$.5$} (P31.45);
    \draw[line width=0.3mm, ->, above] (S21.315) to node[black]{$.5$} (P32.135);
    \draw[line width=0.3mm, ->, left] (S22.270) to node[black]{$.3$} (P33.90);
    \draw[line width=0.3mm, ->, above] (S22.315) to node[black]{$.7$} (P34.135);
    
    \draw[line width=0.3mm, ->]  (P31.270) -- (G41.90);
    \draw[line width=0.3mm, ->]  (P31.225) -- (G40.45);
    \draw[line width=0.3mm, ->]  (P32.315) -- (G43.135);
    \draw[line width=0.3mm, ->]  (P32.270) -- (G42.90);
    \draw[line width=0.3mm, ->]  (P33.225) -- (G43.45);
    \draw[line width=0.3mm, ->]  (P33.270) -- (G44.90);
    \draw[line width=0.3mm, ->]  (P34.270) -- (G45.90);
    \draw[line width=0.3mm, ->]  (P34.315) -- (G46.135);

\end{tikzpicture}
\caption{Example of PC with variables $X_1,\ldots,X_3$. Sum nodes are in blue, product nodes in green, distribution leaf nodes in salmon. In this example, all leaf nodes are univariate. Subscriptions on each $p$ in the figure are used to indicate that those are different leaf distributions (even if sometimes over the same variable).}
\end{figure}
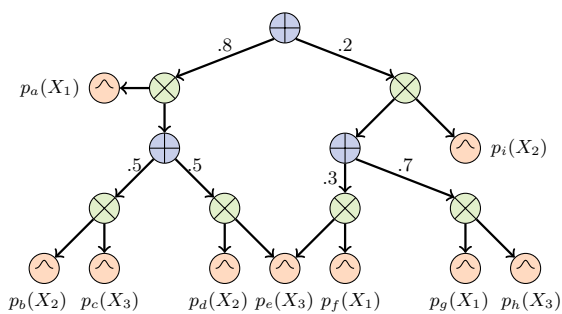

An important feature of normalized valid PCs is their interpretation as hierarchical, discrete mixture models \cite{peharz2016latent,zhao2016unified}: 
\begin{equation}\label{eq:mixture}
    p(\x) = \sum_{z}p(\x | z)p(z) = \sum_{z}p_{z}(\x)p(z) , 
\end{equation}
where $Z$ is a discrete latent vector, which originates from the sum nodes of the structure. The number of states of $Z$, and thus of represented mixture components $p(\x|z)$, grows exponentially in the depth of the PC \cite{peharz2015foundations,zhao2016unified}. While we use this notation here and throughout the paper, we do not run computations directly in this formulation, but instead we make use of the graphical structure of the PC in order to perform efficient tractable inference, as usual for PCs.

\section{Probabilistic circuits with constraints}\label{sec:CPC}

We assume that a normalized valid PC has been produced (learned from data, designed by a human, etc.) over a domain with variables $\X$. Such a PC induces a joint distribution $p(\X)$. The goal is to enforce some (linear) probabilistic propositional logic (PPL) constraints upon $p$. We work with constraints of the form:
\begin{equation}
\sum_{i_{c}}\tau_{i_{c}}\cdot p(F_{i_{c}}) \leq \alpha_{c} , \label{eq:cform}
\end{equation}
where each $F_{i_{c}}$ is a propositional logic formula defined over Boolean variables $\X_{c} = \{X_{j_{c}}\}_{\forall j_c} \subseteq \X$, $\tau_{i_{c}},\alpha_{c}$ are real numbers, $j_c$ (and $i_c$) are indexes of variables (terms) of the constraint $c$, and $c\in C$ is an index over a set of PPL constraints. We assume that constraints are placed in buckets $B$ (mathematically a bucket can be simply an index set indicating the constraints it contains) such that $\X_{B_{1}}\cap\X_{B_{2}}=\emptyset$ for all distinct buckets $B_{1},B_{2}$, where $\X_{B}=\cup_{c\in B}\X_c$ is the union of all variables appearing in a constraint inside bucket $B$. If any $\X_{c_{1}}$ and $\X_{c_{2}}$ of two constraints are not disjoint, then we put them together into the same bucket, so as to ensure that buckets have mutually exclusive sets of variables.

The constraints in each bucket $B$ may obviously create dependencies among the variables $\X_{B}$. In order to avoid inconsistencies between such dependencies and those arising from the graph structure of the PC, we require that the variables in a bucket appear together in nodes of the model, that is, for any $v,B$, $X\in \X_{B}\cap \scope(\node)\Rightarrow \X_{B}\subseteq\scope(\node)$.
Therefore, Equation~\eqref{eq:mixture} can be recast as 
\begin{equation}\label{eq:mixture_ppl}
    p(\X) = \sum_{z}p(z)\prod_{B}p_{z}(\X_{B})\prod_{X_{i} \in \X \setminus \cup \X_{B}}p_{z}(X_{i}), 
\end{equation}
\noindent 
where $p_{z}(\X_{B})$ is a categorical distribution---note that notations $p_{z}(\X_{B})$ and $p_{z}(X_{i})$ employ a slight abuse, as the function itself is ``aware'' of the indexes of the variables in their arguments and may vary accordingly, for example, $p_{z}(X_{i})$ is also a function of $i$ and not only of $X_i$; the same abuse holds elsewhere, for example in Expression~\eqref{eq:cform}. 
Equation~\eqref{eq:mixture_ppl} basically decomposes $p_z(\x)$ of Equation~\eqref{eq:mixture} into components that involve PPL variables (which remain together) and the other variables, which are assumed to be represented by univariate leaf-node distributions. 

\begin{figure}[ht!]
    \centering
    \begin{subfigure}{0.45\textwidth}
        \centering
        \scalebox{0.7}{
\definecolor{blu}{RGB}{199, 206, 234}
\definecolor{gre}{RGB}{181, 234, 215}
\definecolor{re}{RGB}{255, 154, 162}
\definecolor{ore}{RGB}{255, 218, 193}
\definecolor{lgr}{RGB}{226, 240, 203}
\definecolor{mel}{RGB}{255, 183, 178}

\begin{tikzpicture}[
    roundnode/.style={minimum size=7.5mm,
                      inner sep=0}, 
    cross/.style={minimum size=6mm, fill=lgr,
        path picture={
            \draw[black] (path picture bounding box.south east) -- (path picture bounding box.north west) (path picture bounding box.south west) -- (path picture bounding box.north east);
        }
    },
    sum/.style={minimum size=6mm, fill=blu,
        path picture={
            \draw[black] (path picture bounding box.south) -- (path picture bounding box.north) (path picture bounding box.west) -- (path picture bounding box.east);
        }
    },
    gauss/.style={minimum size=6mm, fill=ore,
        path picture={
            \draw[black] plot[domain=-.15:.15] ({\x},{exp(-200*\x*\x -2.)});
        }
    },
    dt/.style={minimum size=6mm, fill=white
    },
    line/.style={
      draw,thick,
      -latex',
      shorten >=2pt
    },
    cloud/.style={
      draw=red,
      thick,
      ellipse,
      fill=red!20,
      minimum height=1em
    },
    indicator/.style={minimum size=6mm, fill=ore,
        path picture={
            \draw[black] plot[domain=-.1:.1] ({\x},{((\x > 0) - 0.5)*0.35});
        }
    }
]

    \node[draw, circle, cross] (p) at (0, 0) {};
    \node[draw, circle, gauss] (spn) at (-1, -1) {};
    \node[draw, circle, gauss] (g) at (1, -1) {};

    \node (ind) at (-1.25, -1.75) {$p(X_{1})$};
    \node (ind) at (1.25, -1.75) {$p(X_{2})$};
    \node (ind) at (0, 0.75) {$p(X_{1}, X_{2})$};

    \draw[line width=0.3mm, ->] (p) to (spn) {};
    \draw[line width=0.3mm, ->] (p) to (g) {};

\end{tikzpicture}}
        \caption{Leaf structure using independence for $X_1$ and $X_2$.}
        \label{fig:b1_fig1}
    \end{subfigure}
    \begin{subfigure}{0.45\textwidth}
        \centering
        \scalebox{0.7}{
\definecolor{blu}{RGB}{199, 206, 234}
\definecolor{gre}{RGB}{181, 234, 215}
\definecolor{re}{RGB}{255, 154, 162}
\definecolor{ore}{RGB}{255, 218, 193}
\definecolor{lgr}{RGB}{226, 240, 203}
\definecolor{mel}{RGB}{255, 183, 178}

\begin{tikzpicture}[
    roundnode/.style={minimum size=7.5mm,
                      inner sep=0}, 
    cross/.style={minimum size=6mm, fill=lgr,
        path picture={
            \draw[black] (path picture bounding box.south east) -- (path picture bounding box.north west) (path picture bounding box.south west) -- (path picture bounding box.north east);
        }
    },
    sum/.style={minimum size=6mm, fill=blu,
        path picture={
            \draw[black] (path picture bounding box.south) -- (path picture bounding box.north) (path picture bounding box.west) -- (path picture bounding box.east);
        }
    },
    gauss/.style={minimum size=6mm, fill=ore,
        path picture={
            \draw[black] plot[domain=-.15:.15] ({\x},{exp(-200*\x*\x -2.)});
        }
    },
    dt/.style={minimum size=6mm, fill=white
    },
    line/.style={
      draw,thick,
      -latex',
      shorten >=2pt
    },
    cloud/.style={
      draw=red,
      thick,
      ellipse,
      fill=red!20,
      minimum height=1em
    },
    indicator/.style={minimum size=6mm, fill=ore,
        path picture={
            \draw[black] plot[domain=-.1:.1] ({\x},{((\x > 0) - 0.5)*0.35});
        }
    }
]

    \node[draw, circle, sum, label=above:{$p(X_{1}, X_{2})$}] (s1) at (0, 0.5) {};
    \node[draw, circle, cross] (p1) at (-1, -1) {};
    \node[draw, circle, cross] (p2) at (1, -1) {};
    \node[draw, circle, indicator, label=below:{$x_1$}] (g1) at (-1.5, -2.5) {};
    \node[draw, circle, indicator, label=below:{$\lnot x_2$}] (g2) at (-0.5, -2.5) {};
    \node[draw, circle, indicator, label=below:{$\lnot x_1$}] (g3) at (0.5, -2.5) {};
    \node[draw, circle, indicator, label=below:{$x_2$}] (g4) at (1.5, -2.5) {};

    \node[draw, circle, cross] (p11) at (-3, -1) {};
    \node[draw, circle, cross] (p21) at (3, -1) {};
    \node[draw, circle, indicator, label=below:{$x_1$}] (g11) at (-3.5, -2.5) {};
    \node[draw, circle, indicator, label=below:{$x_2$}] (g21) at (-2.5, -2.5) {};
    \node[draw, circle, indicator, label=below:{$\lnot x_1$}] (g31) at (2.5, -2.5) {};
    \node[draw, circle, indicator, label=below:{$\lnot x_2$}] (g41) at (3.5, -2.5) {};

    \draw[line width=0.3mm, ->] (s1) to node[black, xshift=-0.35cm]{$\theta_{10}$} (p1);
    \draw[line width=0.3mm, ->] (s1) to node[black, xshift=-0.35cm]{$\theta_{01}$} (p2);
    \draw[line width=0.3mm, ->] (p11) to node[black, yshift=0.25cm, xshift=-0.25cm]{} (g11);
    \draw[line width=0.3mm, ->] (p11) to node[black, yshift=0.25cm, xshift=0.25cm]{} (g21);
    \draw[line width=0.3mm, ->] (p2) to node[black, yshift=0.25cm, xshift=-0.25cm]{} (g3);
    \draw[line width=0.3mm, ->] (p2) to node[black, yshift=0.25cm, xshift=0.25cm]{} (g4);

    \draw[line width=0.3mm, ->] (s1) to node[black, xshift=-0.35cm]{$\theta_{11}$} (p11);
    \draw[line width=0.3mm, ->] (s1) to node[black, xshift=-0.35cm]{$\theta_{00}$} (p21);
    \draw[line width=0.3mm, ->] (p1) to node[black, yshift=0.25cm, xshift=-0.25cm]{} (g1);
    \draw[line width=0.3mm, ->] (p1) to node[black, yshift=0.25cm, xshift=0.25cm]{} (g2);
    \draw[line width=0.3mm, ->] (p21) to node[black, yshift=0.25cm, xshift=-0.25cm]{} (g31);
    \draw[line width=0.3mm, ->] (p21) to node[black, yshift=0.25cm, xshift=0.25cm]{} (g41);

\end{tikzpicture}}
        \caption{Leaf replacement structure without assuming independence of $X_1$ and $X_2$.}
        \label{fig:b1_fig2}
    \end{subfigure}
    \caption{Leaf distribution replacement structures that can be used to represent the parameters of a categorical variable for a bucket $B$ with $\X_{B} = \{X_{1}, X_{2}\}$.}
    \label{fig:decomp}
\end{figure}
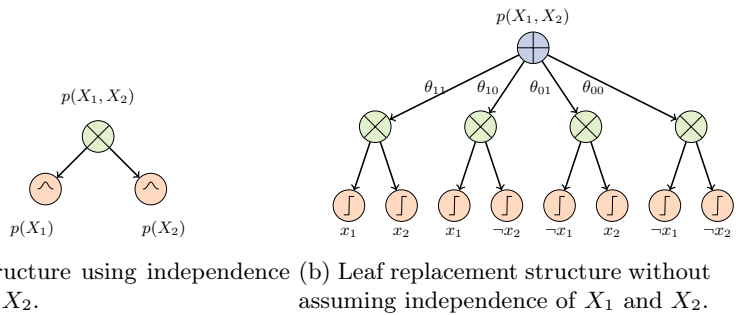

For ease of exposure, but also for the sake of compatibility with software that only deals with univariate leaf distributions, one can replace categorical distributions in leaf nodes with new sub-PCs. If one assumes independence among scope variables, then a product node with univariate leaf nodes suffices. If one wants to fully exploit the categorical distribution node, then a sum node with one child per parameter of the categorical distribution can be used. Figure~\ref{fig:decomp} gives an example of dealing with a categorical ``joint'' distribution over two Boolean variables $X_1,X_2$. Figure~\ref{fig:b1_fig1} shows the independent case, while Figure~\ref{fig:b1_fig2} shows the joint approach to represent the distribution for $X_1$ and $X_2$. The reader may have already noticed that large buckets of constraints will force the model to keep together many variables, which can be problematic as the number of parameters of the categorical joint distribution of all variables in a bucket $B$ will grow exponentially in $|\X_B|$ (as in Figure~\ref{fig:b1_fig2}, all possible configurations of $\X_B$ would be listed). We will discuss this later, and ask the reader to assume that buckets (or equivalently scopes of leaf distribution nodes) are not large.

Given a PC representing $p(\X)$ and PPL constraints, we aim to find an \emph{efficient} approach to discover a new PC inducing a distribution $q^*(\X)$ that is close to $p(\X)$ while respecting the PPL constraints: 
\begin{equation}\label{eq:gen_opt0}
    \begin{split}
        q^{*}(\X) & = \argmin_{q(\X)} \mathcal{L}(p(\X), q(\X)) \\
        \text{s.t.} & \quad
            \forall B, \forall c\in B: \sum_{i_{c}}\tau_{i_{c}}\cdot q(F_{i_{c}}) \leq \alpha_{c}', ,\qquad 
            q(\X) \in \mathcal{P}(\X)\, ,
    \end{split}
\end{equation}
\noindent where $\mathcal{L}$ measures the discrepancy between two distributions, 
$\mathcal{P}(\X)$ denotes a set of probability distributions over $\X$ that can be represented by a PC, and $B$ are buckets of PPL constraints. Optimization~\eqref{eq:gen_opt0} is impractical, as it amounts to solving a complex optimization problem to search over $\mathcal{P}(\X)$, even if $\mathcal{L}$ is simple enough. Therefore, we exploit the PC on which $p(\X)$ was estimated, in order to constraint the search space of $q(\X)$: we enforce $q(\X)$ to have a shape similar to $p(\X)$, i.e.
\begin{equation}\label{eq:mixture_q}
    q(\X) = \sum_{z}p(z)\prod_{B}q_{z}(\X_{B})\prod_{X_{i} \in \X \setminus \cup \X_{B}}p_{z}(X_{i}),
\end{equation}
\noindent that is, only $\prod_{B}q_{z}(\X_{B})$ will differ from the specification of $p(\X)$. Plainly put, we only refine the distributions in the leaf nodes of the PC. Moreover, we use the Kullback-Leibler divergence
$\mathcal{L}(p(\X),q(\X)) = H(p(\X),q(\X)) - H(p(\X))$ as discrepancy measure, where $H(\cdot)$ is the entropy and $H(\cdot,\cdot)$ the cross entropy. 
Clearly, we can focus on the cross entropy only, as the second term does not contain $q(\X)$. Our first result is an upper bound on the cross entropy which allows us to run the optimization efficiently. The bound on the cross-entropy establishes an upper bound on the KL-divergence between $p(\X)$ and $q(\X)$. 

\begin{theorem}\label{th:bound}
Assume a PC representing a distribution $p(\X)$ as in Equation~\eqref{eq:mixture_ppl} and PPL constraints as in Equation~\eqref{eq:cform} (placed in disjoint buckets $B$) are given. Assume that $q(\X)$ is a distribution induced by a PC with form as in Equation~\eqref{eq:mixture_q}. Then, 
$H(p(\X),q(\X)) \leq \sum_{B}\mathbb{E}_{z}[H(p_{z}(\X_{B}), q_{z}(\X_{B}))] + H(p(\X',Z))$, 
where $\X'$ are the variables not appearing in constraints.
\end{theorem}
\begin{proof}
Note that we are particularly interested in terms with parameters in $q(\X)$, as they will be optimized later. First, recall that 
\begin{equation}\label{eq:apndx1}
    \begin{split}
        &-H(p(\X), q(\X)) = \sum_{\x}p(\x)\log q(\x)\, ,
    \end{split}
\end{equation}
\noindent
and for any configuration $\x$ of $\X$ and for any arbitrary $z_0\in Z$, we have:
\begin{equation}\label{eq:apndx2}
    q(\x) = \sum_{z}p(z)p'_{z}(\x')\prod_{B}q_{z}(\x_{B}) \geq p(z_0)p'_{z_0}(\x')\prod_{B}q_{z_0}(\x_{B})\, ,
\end{equation}
\noindent
which holds because all terms are non-negative,
where $\X' = \X \setminus \cup \X_{B}$ (variables not in any constraint), and $p'_{z}(\X') = \prod_{X_{i} \in \X'}p_{z}(X_{i})$, for given $z$. By substituting \eqref{eq:apndx2} into \eqref{eq:apndx1}, we can establish a lower bound on the negative cross-entropy term:
\begin{align}
- H(p(\X), q(\X))& \geq \sum_{\x}p(\x)\log[p(z_0)p'_{z_0}(\x')\prod_{B}q_{z_0}(\x_{B})]\,\label{eq:apndx4a} \\
& = \sum_{\x}\sum_{z}p(z)p'_{z}(\x')\prod_{\beta}p_{z}(\x_{\beta})\log[p(z)p'_{z}(\x')\prod_{B}q_{z}(\x_{B})],
\label{eq:apndx4}
\end{align}
\noindent
with the arbitrary $z_0\in Z$ in Expression~\eqref{eq:apndx4a} being chosen to be equal to $z$ for each of the elements in the summation over $z$, thus resulting in Expression~\eqref{eq:apndx4}. 
Then, we can split Expression~\eqref{eq:apndx4} into two parts (using the log of products as sum of logs), where only the second term depends on $q(\X)$: 
\begin{multline}
- H(p(\X), q(\X)) \geq \sum_{\x}\sum_{z}p(z)p'_{z}(\x')\prod_{\beta}p_{z}(\x_{\beta})\log[p(z)p'_{z}(\x')] \\ +
\sum_{\x}\sum_{z}p(z)p'_{z}(\x')\prod_{\beta}p_{z}(\x_{\beta})\log[\prod_{B}q_{z}(\x_{B})] . \label{eq:apndx44}
\end{multline}
\noindent The first term in the RHS of Expression~\eqref{eq:apndx44} can be reduced to $-H(p(\X',Z))$; it does not depend on $q(\X)$, and will consequently not be analyzed further. The second term in the RHS can be manipulated as 
\begin{align}
        &= \sum_B\sum_{\x}\sum_{z}p(z)p'_{z}(\x')\prod_{\beta}p_{z}(\x_{\beta})\log q_{z}(\x_{B}) \nonumber \\
        &=\sum_B \sum_{\substack{\x_{B_t}\\ \forall t}}\sum_{\x'} \sum_{z}p(z)p'_{z}(\x')\prod_{\beta}p_{z}(\x_{\beta})\log q_{z}(\x_{B}) \nonumber \\
        &= \sum_{B}\sum_{z}p(z)\sum_{\substack{\x_{B_t}\\ \forall t}}\prod_{\beta}p_{z}(\x_{\beta})\log q_{z}(\x_{B})\sum_{\x'}p'_{z}(\x') \nonumber \\
        &= \sum_{B}\sum_{z}p(z)\sum_{\substack{\x_{B_t}\\ \forall t}}\prod_{\beta}p_{z}(\x_{\beta})\log q_{z}(\x_{B}) \nonumber \\
        &= \sum_{B}\sum_{z}p(z)\sum_{\x_{B}}
        p_{z}(\x_{B})\log q_{z}(\x_{B})
        \sum_{\substack{\x_{B_t}\\ \forall t, B_t\neq B}}\prod_{\beta\neq B}p_{z}(\x_{\beta}) \nonumber \\
        &= \sum_{B}\sum_{z}p(z)\sum_{\x_{B}}
        p_{z}(\x_{B})\log q_{z}(\x_{B})
        \prod_{\beta\neq B} \sum_{\x_{\beta}} p_{z}(\x_{\beta}) \nonumber \\
        &= \sum_{B}\sum_{z}p(z)\sum_{\x_{B}}
        p_{z}(\x_{B})\log q_{z}(\x_{B}) \nonumber \\
        &= -\sum_{B}\mathbb{E}_{z}[ H(p_{z}(\X_{B}), q_{z}(\X_{B}))] \label{eq:apndx5} .
\end{align}
\noindent Hence, $- H(p(\X), q(\X)) \geq -H(p(\X',Z)) -\sum_{B}\mathbb{E}_{z}[ H(p_{z}(\X_{B}), q_{z}(\X_{B}))]$, and the result follows. \qed
\end{proof}

Thus, we can adapt the PC at hand using the specified constraints by minimizing the upper bound on the desired discrepancy, leaving aside the term $H(p(\X',Z))$ which does not involve $q(\X)$:
\begin{equation}\label{eq:gen_opt0bound}
    \begin{split}
        q^{*}(\X) & = \argmin_{q(\X)} \sum_{B}\mathbb{E}_{z}[H(p_{z}(\X_{B}), q_{z}(\X_{B}))] \\
        \text{s.t.} & \quad
            \forall B, \forall c\in B: \sum_{i_{c}}\tau_{i_{c}}\cdot q(F_{i_{c}}) \leq \alpha_{c}\, ,\qquad
            q(\X) \in \mathcal{P}(\X)\, ,
    \end{split}
\end{equation}
Theorem~\ref{th:optcpl} sheds light on the complexity of the procedure; it is based on considerably mild assumptions, as long as buckets do not involve too many variables. 

\begin{theorem}\label{th:optcpl}
Given the same inputs as Theorem~\ref{th:bound}, and assuming $|\X_B|\leq k$ for all buckets $B$, the solution $q^*$ to the optimization in Optimization~\eqref{eq:gen_opt0bound} can be found in polynomial time in the input size (while possibly exponential in $k$). 
\end{theorem}
\begin{proof}
The objective function is a sum over buckets containing (mutually) disjoint sets of variables, so we can solve Optimization~\eqref{eq:gen_opt0bound} by solving separate optimizations for each bucket $B$:
\begin{equation}\label{eq:gen_opt_dec}
    \begin{split}
\forall B:\quad  q^{*}(\X_B) & = \argmin_{q(\X_B)} -\sum_z p(z)\sum_{\x_B} p_z(\x_B) \log q_z(\x_B) \\
        \text{s.t.} & \quad
            \forall c\in B: \sum_{i_{c}}\tau_{i_{c}}\cdot q(F_{i_{c}}) \leq \alpha_{c}\, ,\qquad
            q(\X_B) \in \mathcal{P}(\X_B)\, .
    \end{split}
\end{equation}

\noindent (Note the abuse of notation here, as $q^*(\X_B)$ is used to indicate the parameters of model $q^*(\X)$ that are associated with leaf nodes containing variables $\X_B$.) Optimization~\eqref{eq:gen_opt_dec} can be solved for each $B$ using convex optimization solvers, which run in polynomial time in the size of their inputs (and can be very efficient in practice). The values $p_z(\x_B)$ and $p(z)$ are fixed during the optimization and can be obtained directly from the PC model representing $p(\X)$. 

Assuming that $q_z(\x_B)$ is parameterized using values $\theta_{z,\x_B}$ representing a categorical distribution over $\X_B$ conditional to $Z=z$ (same structure as in Figure \ref{fig:b1_fig2}), we obtain Optimization~\eqref{eq:gen_opt_dec2} for each bucket $B$. 
Note that in Optimization~\eqref{eq:gen_opt_dec2}, each PPL formula $F_{i_c}$ is written down as the sum of the worlds that satisfy the formula (we can query $F_{i_{c}}(\x_B)$ to see if each $\x_B$ satisfies $F_{i_c}$, assuming $F_{i_c}=1$ if so, and zero otherwise). Optimization~\eqref{eq:gen_opt_dec2} also connects the local parameters $\theta_{z,\x_B}$ with the marginal value of the candidate PC for $\x_B$, that is, $q(\x_B)=\sum_z p(z) \theta_{z,\x_B}$, which appears in the last expression of the optimization problem: thus, the imposed constraint is a global constraint in the joint model $q(\X)$, and not simply a local constraint in the local parameters. Note also that $p(z)=q(z)$ (by assumption from Expression~\eqref{eq:mixture_q}). 
\begin{equation}\label{eq:gen_opt_dec2}
    \begin{split}
        \forall B:\quad & q^{*}(\X_B) = \argmin_{\theta_{z,\x_B}:~\forall z,\x_B} -\sum_z p(z)\sum_{\x_B} p_z(\x_B) \log \theta_{z,\x_B}\, , \\
        & \quad\text{s.t.} \quad
        \begin{split}
        &\forall z,\x_B:~ \theta_{z,\x_B}\geq 0\, ,\qquad \forall z:~ 1 = \sum_{\x_B} \theta_{z,\x_B}\, ,\\
            &\forall c\in B: \sum_{i_{c}}\tau_{i_{c}}\cdot \left(\sum_{\x_B} F_{i_{c}}(\x_B) (\sum_z p(z) \theta_{z,\x_B})\right) \leq \alpha_{c}\, . 
    \end{split}
    \end{split}
\end{equation}
\noindent
Optimization~\eqref{eq:gen_opt_dec2}, and hence Optimizations~\eqref{eq:gen_opt0} and~\eqref{eq:gen_opt_dec}, will have a feasible solution so long as the set of PPL constraints has a feasible solution. This can be checked using linear programming using the constraints in Optimization~\eqref{eq:gen_opt_dec2}. Therefore, it can be checked in polynomial time if the user provided an infeasible set of constraints. The number of buckets is bounded by the number of constraints $C$, which therefore also bounds the number of optimization calls. The optimization for bucket $B$ has $O(|Z|\cdot |\X_B|)$ variables and $O(C\cdot |\X_B|)$ constraints (those are all very loose bounds), which is asymptotically bounded by the PC size plus constraints' size (that is, the input size), and convex optimization can be solved in polynomial time in the number of variables and constraints.\qed
\end{proof}

\section{Experiments}\label{sec:exp}

\begin{figure*}[thp!]
    \centering
    \begin{subfigure}[b]{0.32\textwidth}
        \centering
        \includegraphics[width=\linewidth]{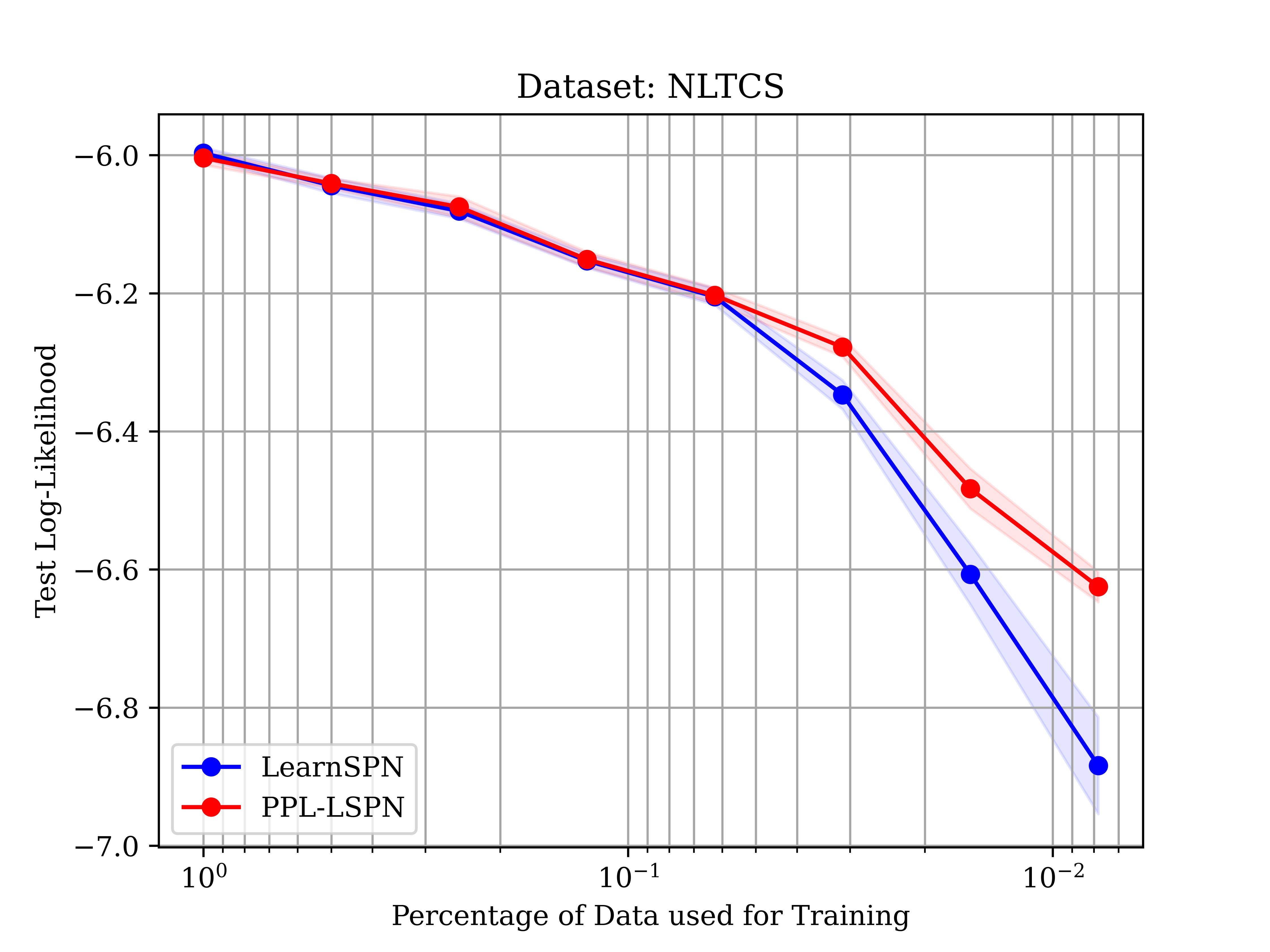}
    \end{subfigure}
    \begin{subfigure}[b]{0.32\textwidth}
        \centering
        \includegraphics[width=\linewidth]{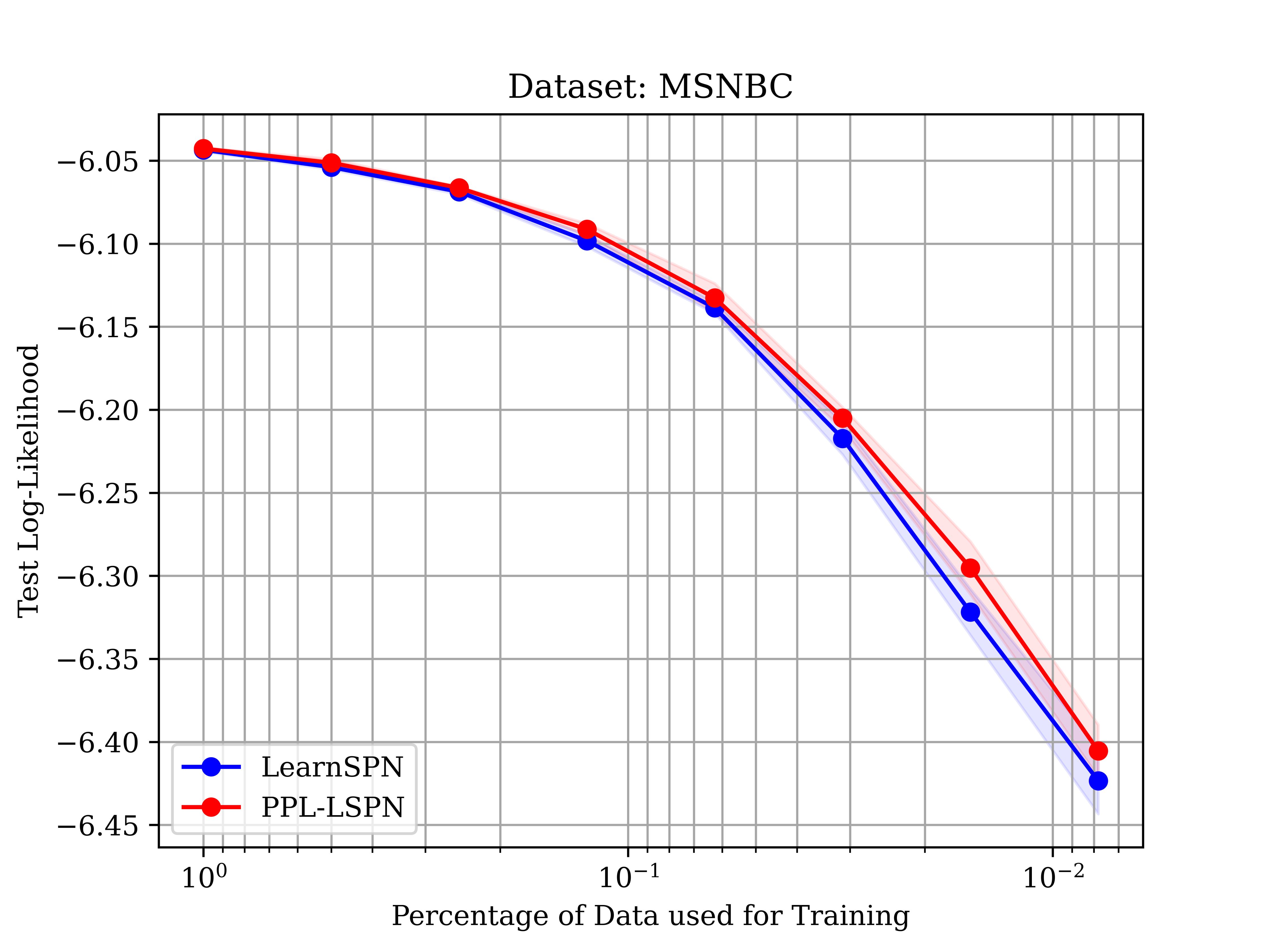}
    \end{subfigure}
    \begin{subfigure}[b]{0.32\textwidth}
        \centering
        \includegraphics[width=\linewidth]{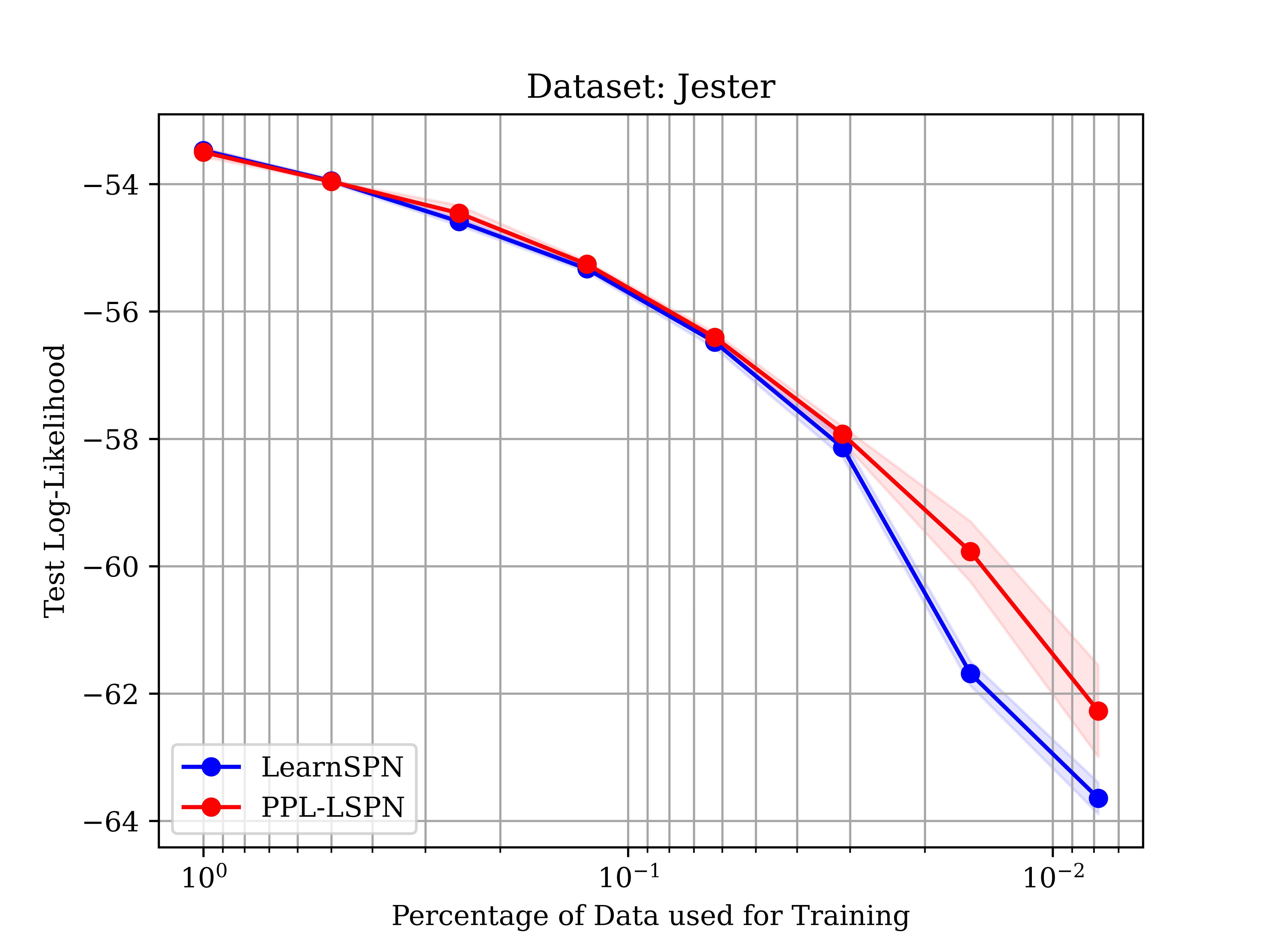}
    \end{subfigure}
    \caption{LearnSPN vs. (constrained) PPL-LSPN trained on scarce datasets.}
    \label{fig:scrc-lspn}
\end{figure*}

\begin{figure*}[th!]
     \centering
     \begin{subfigure}[b]{0.32\textwidth}
         \centering
         \includegraphics[width=\linewidth]{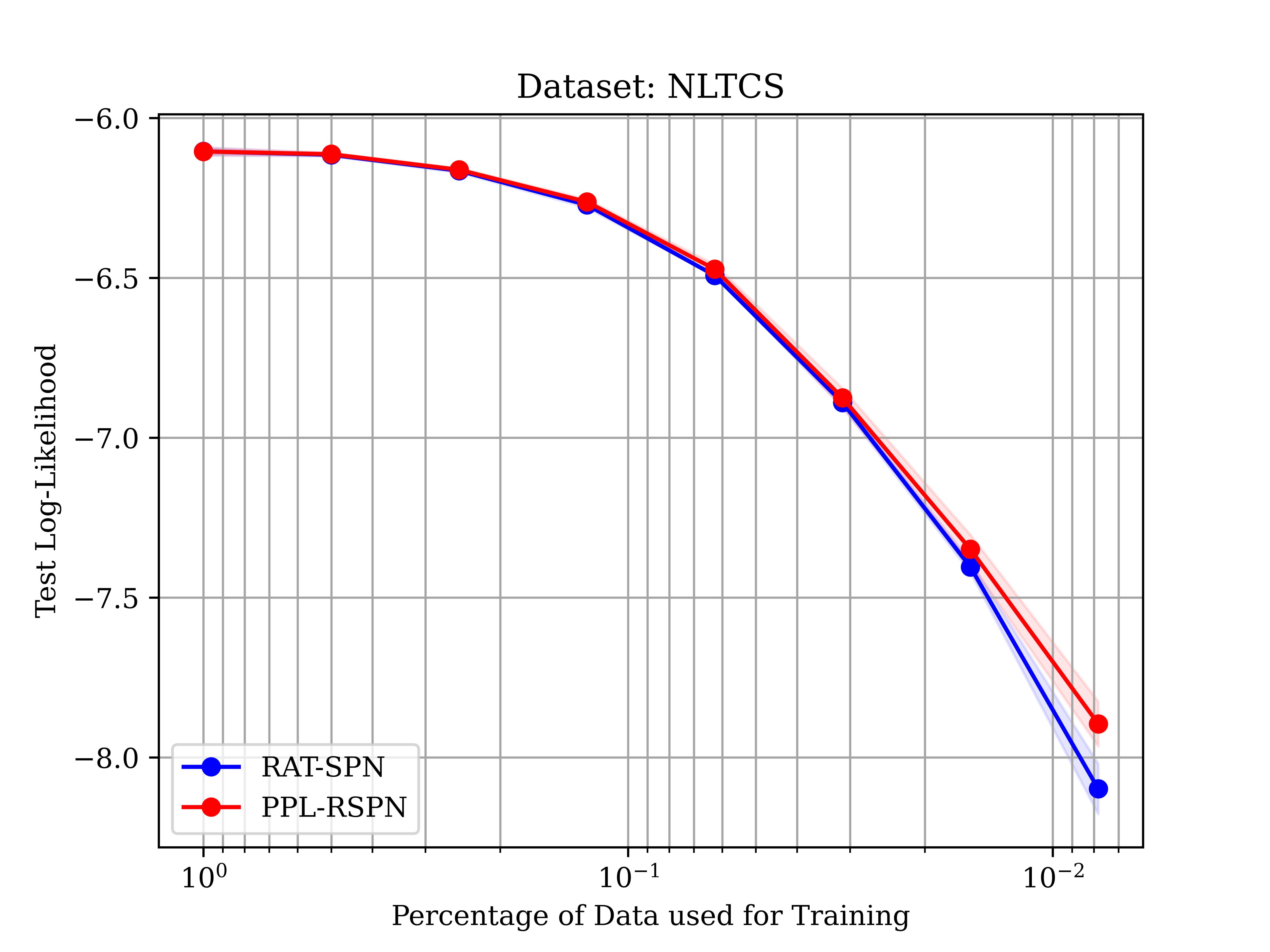}
     \end{subfigure}
     \begin{subfigure}[b]{0.32\textwidth}
         \centering
         \includegraphics[width=\linewidth]{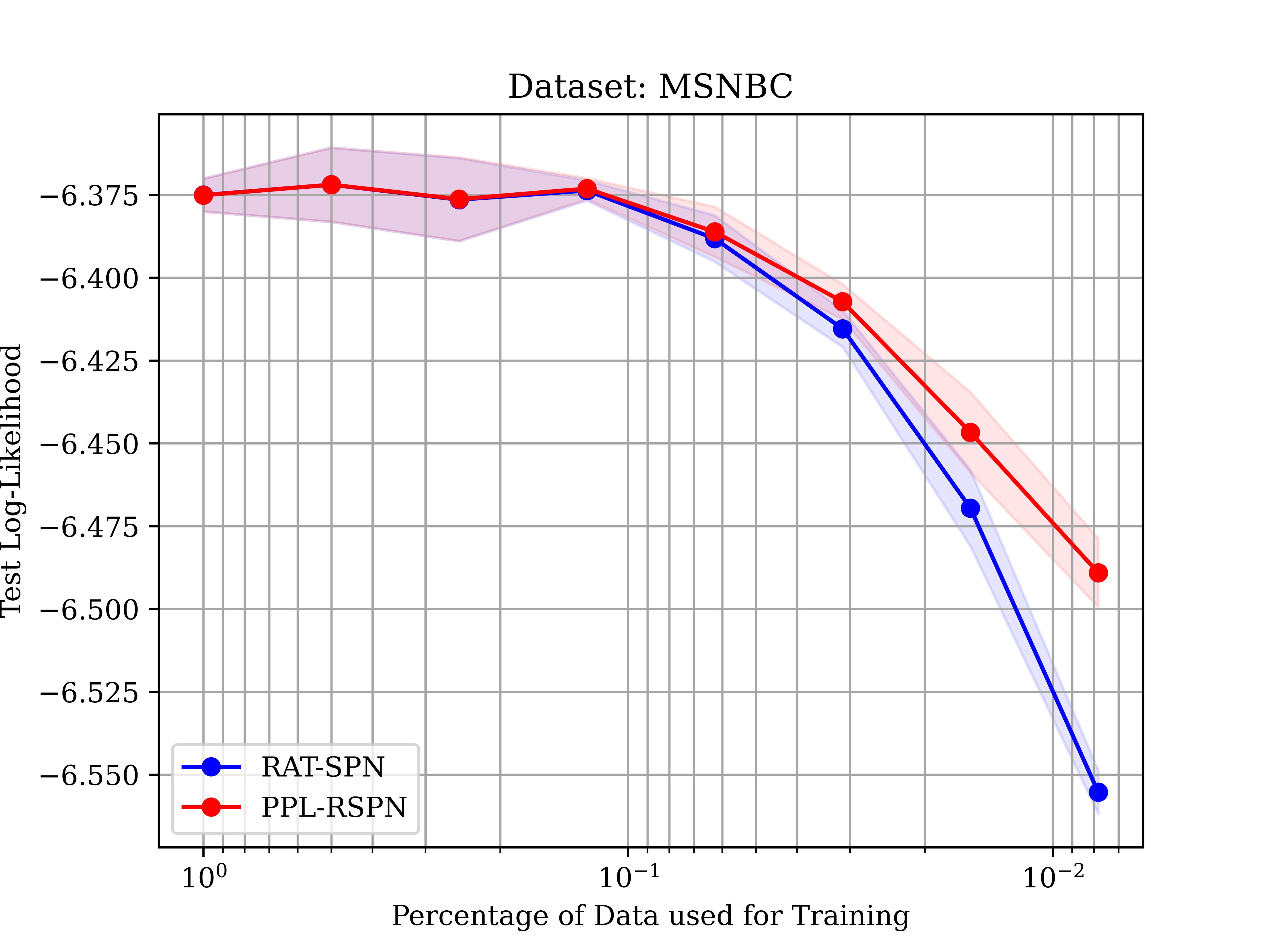}
     \end{subfigure}
     \begin{subfigure}[b]{0.32\textwidth}
         \centering
         \includegraphics[width=\linewidth]{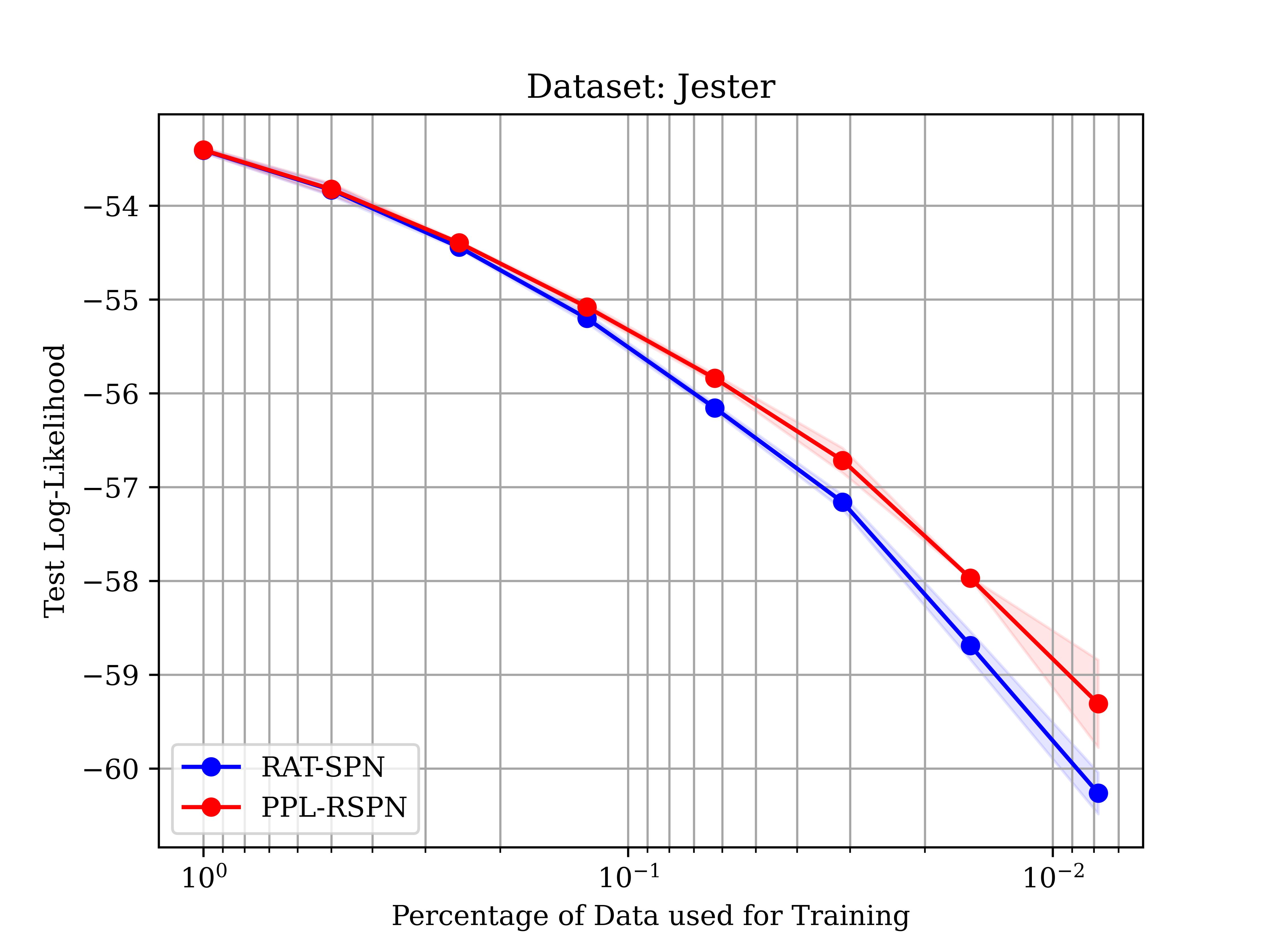}
     \end{subfigure}
     \caption{RAT-SPN vs. (constrained) PPL-RSPN trained on scarce datasets.}
     \label{fig:scrc_rspn}
 \end{figure*}

 \begin{figure*}[th!]
     \centering
     \begin{subfigure}[b]{0.32\textwidth}
         \centering
         \includegraphics[width=\linewidth]{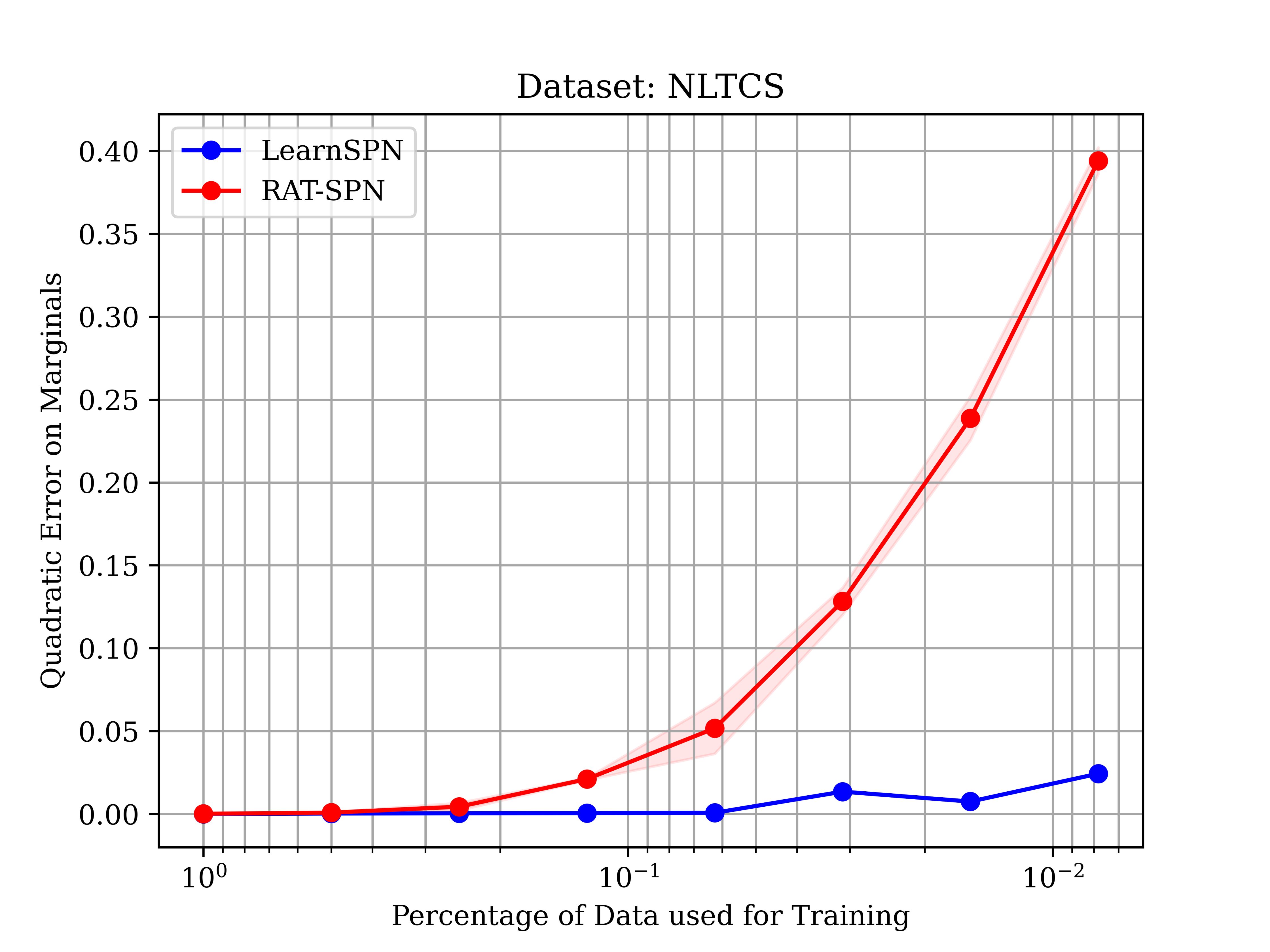}
     \end{subfigure}
     \begin{subfigure}[b]{0.32\textwidth}
         \centering
         \includegraphics[width=\linewidth]{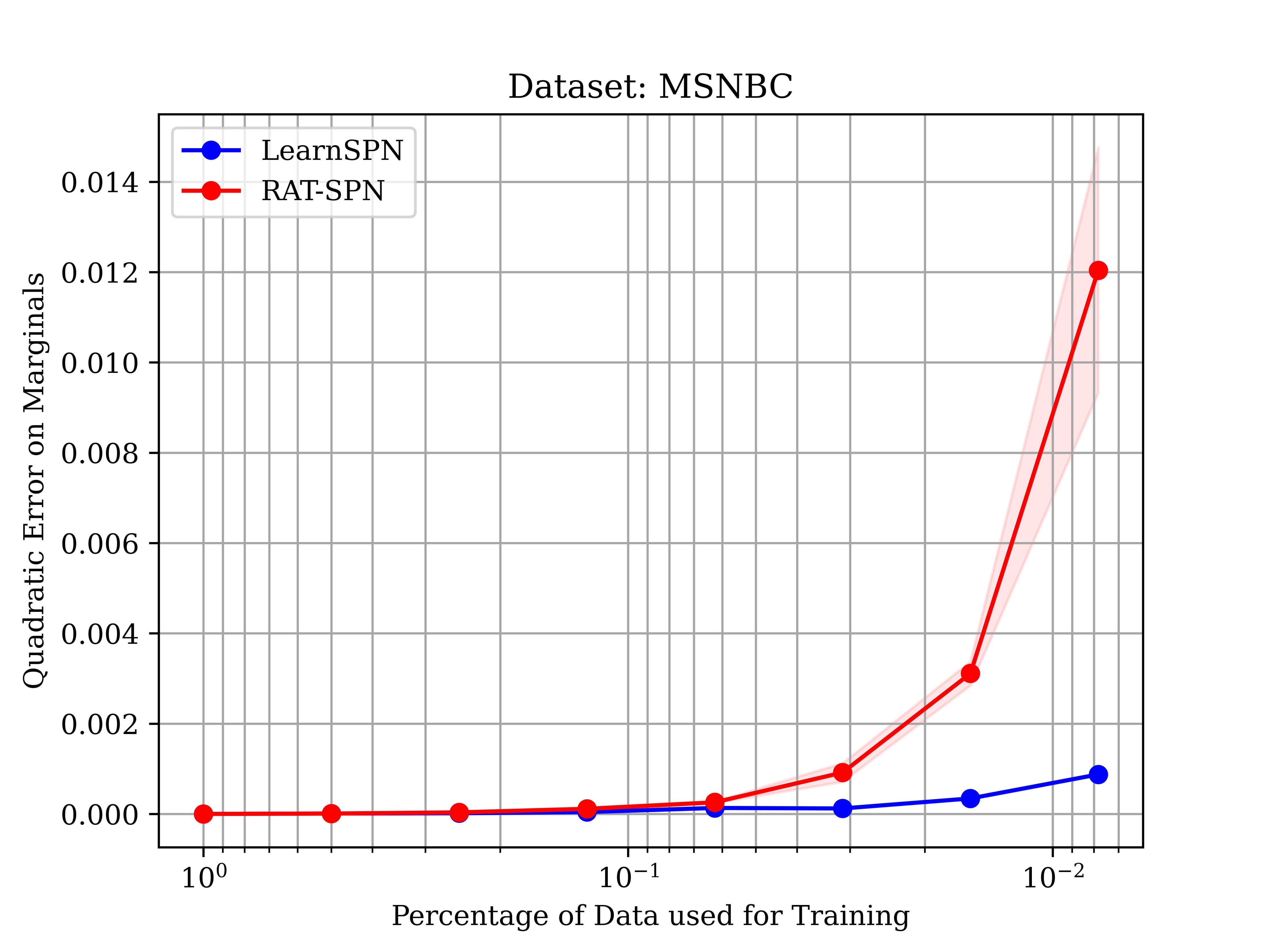}
     \end{subfigure}
     \begin{subfigure}[b]{0.32\textwidth}
         \centering
         \includegraphics[width=\linewidth]{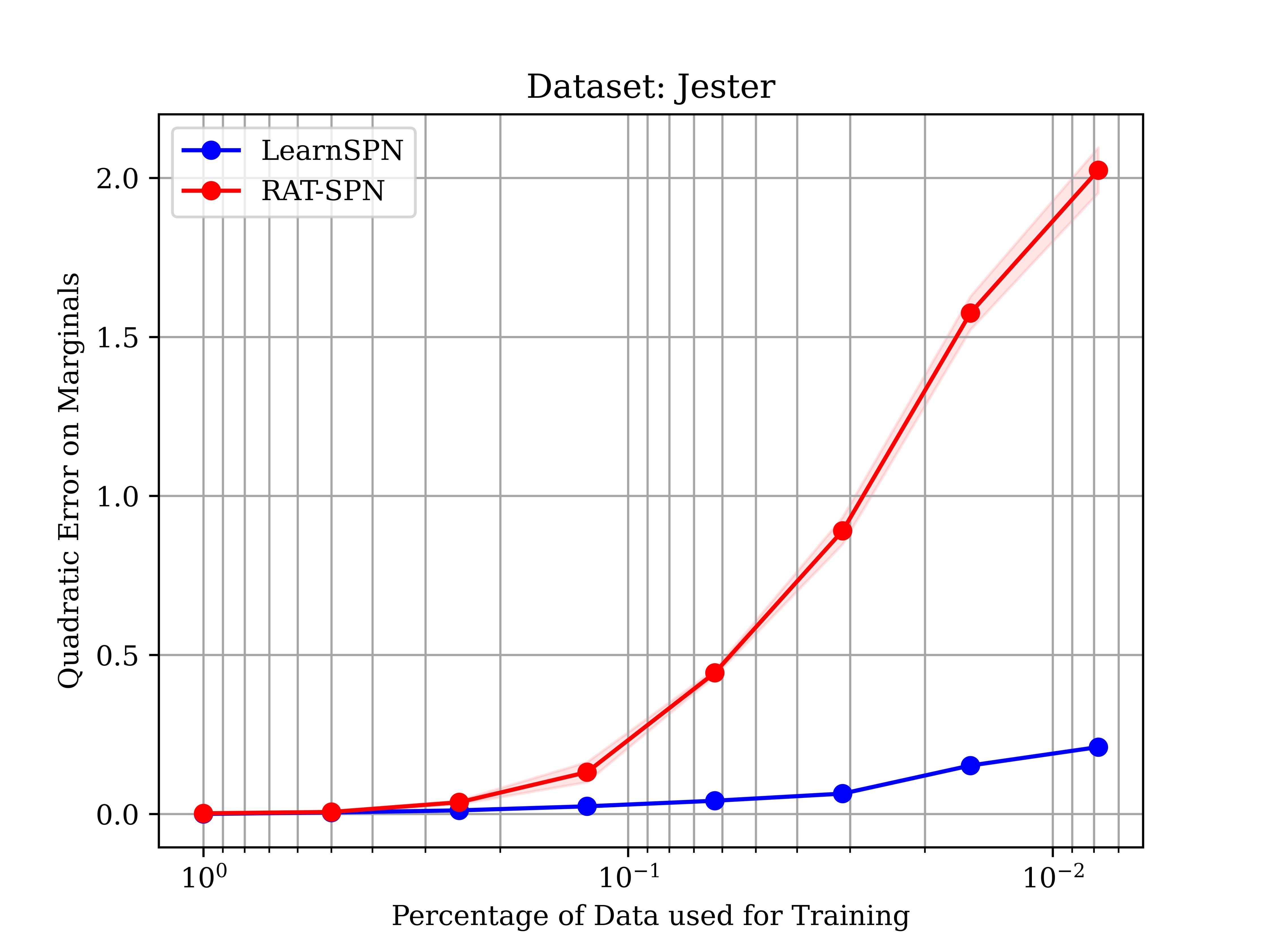}
     \end{subfigure}
     \caption{Sum of quadratic differences on marginal parameters between the models with and without marginal constraints, when trained on scarce data. Constraints clearly refine the model more strongly for RAT-SPNs than for LearnSPN. Standard RAT-SPN marginals are very far from matching the empirical marginal distributions (data not shown).}
     \label{fig:quad_err}
 \end{figure*}

We conduct a series of experiments to illustrate how constrained optimization can be utilized to shape a desirable performance or behavior in PCs. For the sake of this illustration, we focus on two use cases of constraints, namely (i) constraints over marginals of the distribution, and (ii) constraints for enforcing fairness in distributions.

The idea behind constraints on marginals is to adjust a probabilistic model to match the empirical marginals of $\X$ on data $\mathcal{D}$. Typically, it is easier to accurately learn the marginal distributions over single variables rather than the whole joint distribution, in particular in cases when $\mathcal{D}$ is scarce and/or incomplete. In Sections \ref{exp_scrc} and \ref{exp_mv} we explore how the use of constraints on empirical marginals affects the performance/behavior of learned PCs.

In Section \ref{exp_fair}, we investigate the impact of applying our method to a variety of fairness-specific classification tasks by adding fairness in the form of PPL constraints into PCs. Most common PC learning methods are not known to be inherently compatible with fairness, and being able to apply fairness constraints to PCs opens the door to utilizing these probabilistic models in areas where fairness is a priority, thus extending their domain of applicability. 

Throughout the experiments, we utilize both LearnSPN \cite{gens2013learning} and RAT-SPN \cite{peharz2020random} for learning baseline PC models (one could also handcraft a PC for a purpose and use it with our approach, as we are not bound by the way the PC was obtained). We use the original implementation of RAT-SPN\footnote{\hyperlink{https://github.com/cambridge-mlg/RAT-SPN}{https://github.com/cambridge-mlg/RAT-SPN}}, and the implementation of \cite{correia2020joints}\footnote{\hyperlink{https://github.com/AlCorreia/GeFs}{https://github.com/AlCorreia/GeFs}} for LearnSPN. For LearnSPN, statistical test significance and the Laplace smoothing parameter are set to 0.01 over all experiments. For RAT-SPN, hyperparameters that correspond to the region graph structure are set as follows: the number of recursive splits is 10, the depth of each recursive split is 2, the number of input distributions in each partition is 8, and the number of sum nodes per partition is 8; all the other hyperparameters are set to their defaults. For each experiment, RAT-SPN is trained for 20 epochs.

 \begin{figure*}[htp!]
     \centering
     \begin{subfigure}[b]{0.32\textwidth}
         \centering
         \includegraphics[width=\linewidth]{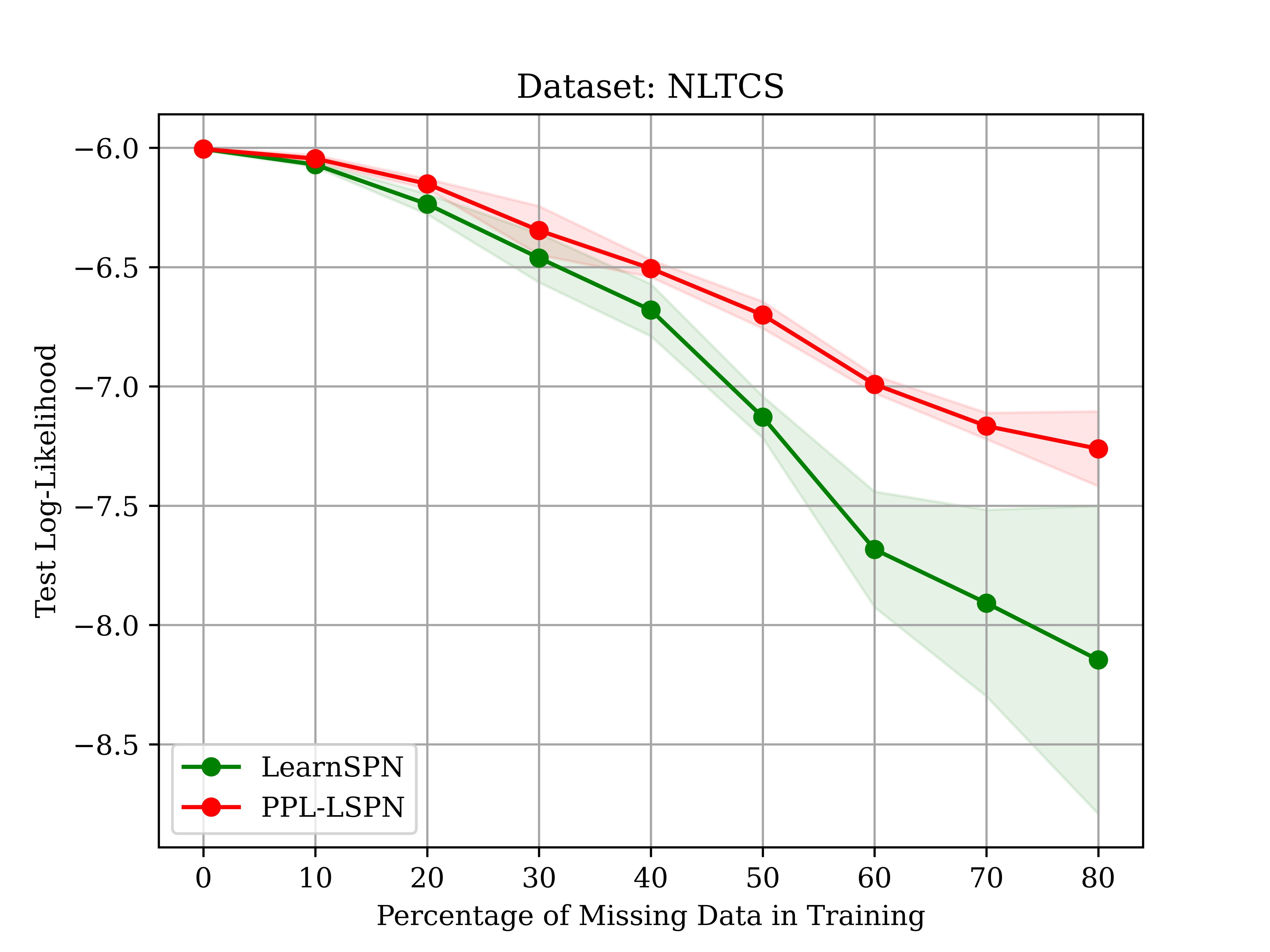}
    \end{subfigure}
     \begin{subfigure}[b]{0.32\textwidth}
         \centering
         \includegraphics[width=\linewidth]{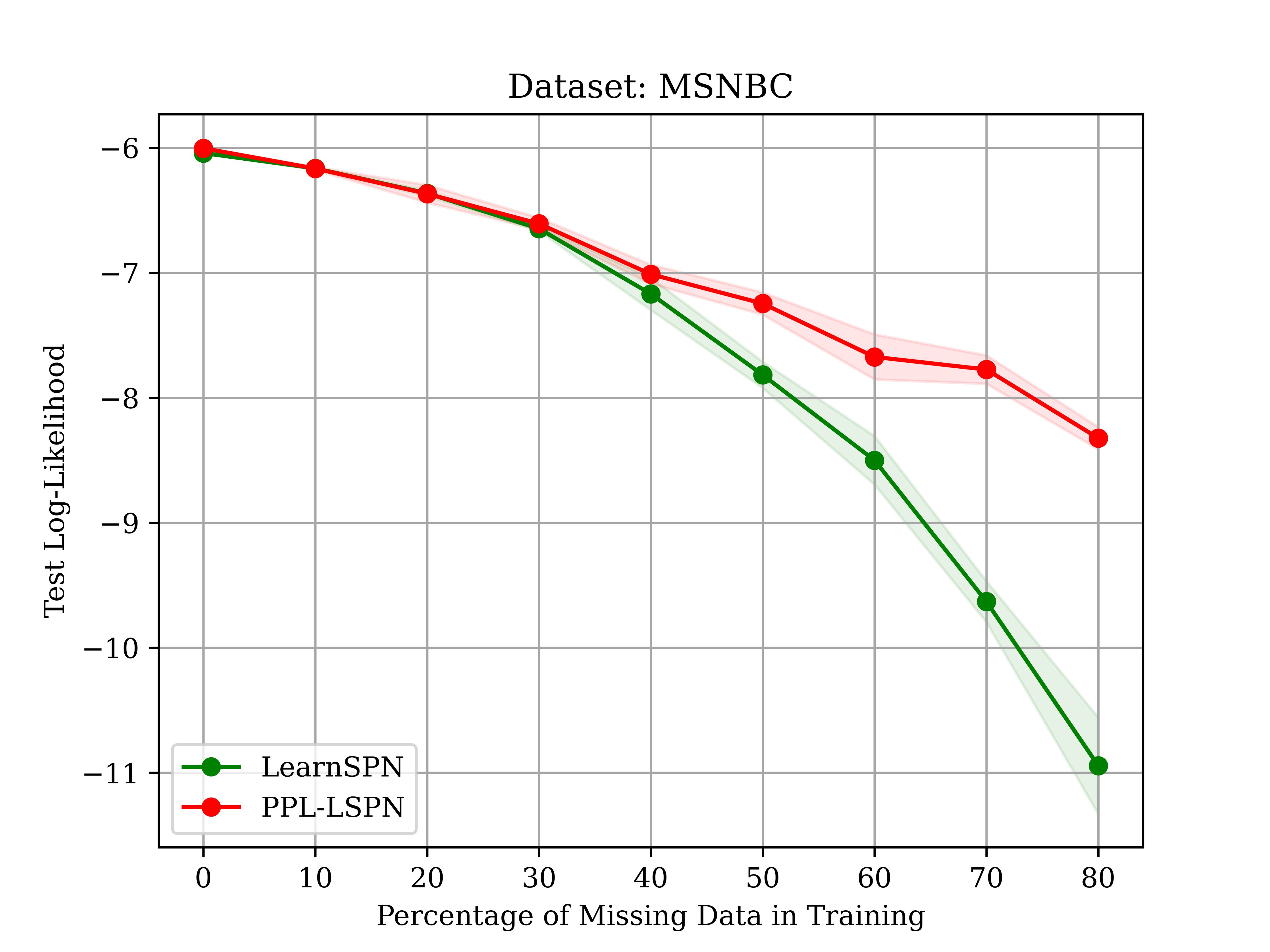}
     \end{subfigure}
     \begin{subfigure}[b]{0.32\textwidth}
         \centering
         \includegraphics[width=\linewidth]{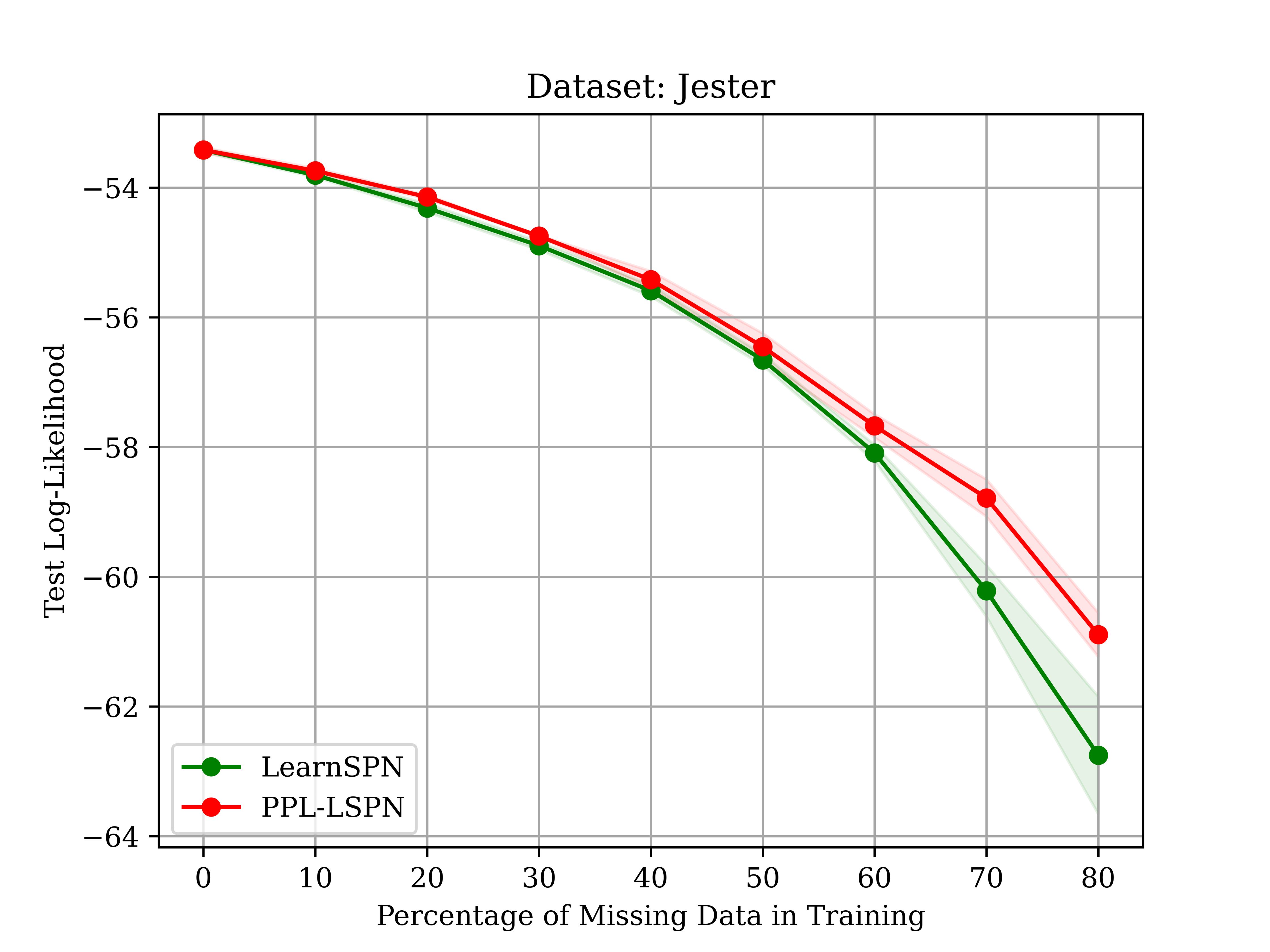}
     \end{subfigure}
     \begin{subfigure}[b]{0.32\textwidth}
         \centering
         \includegraphics[width=\linewidth]{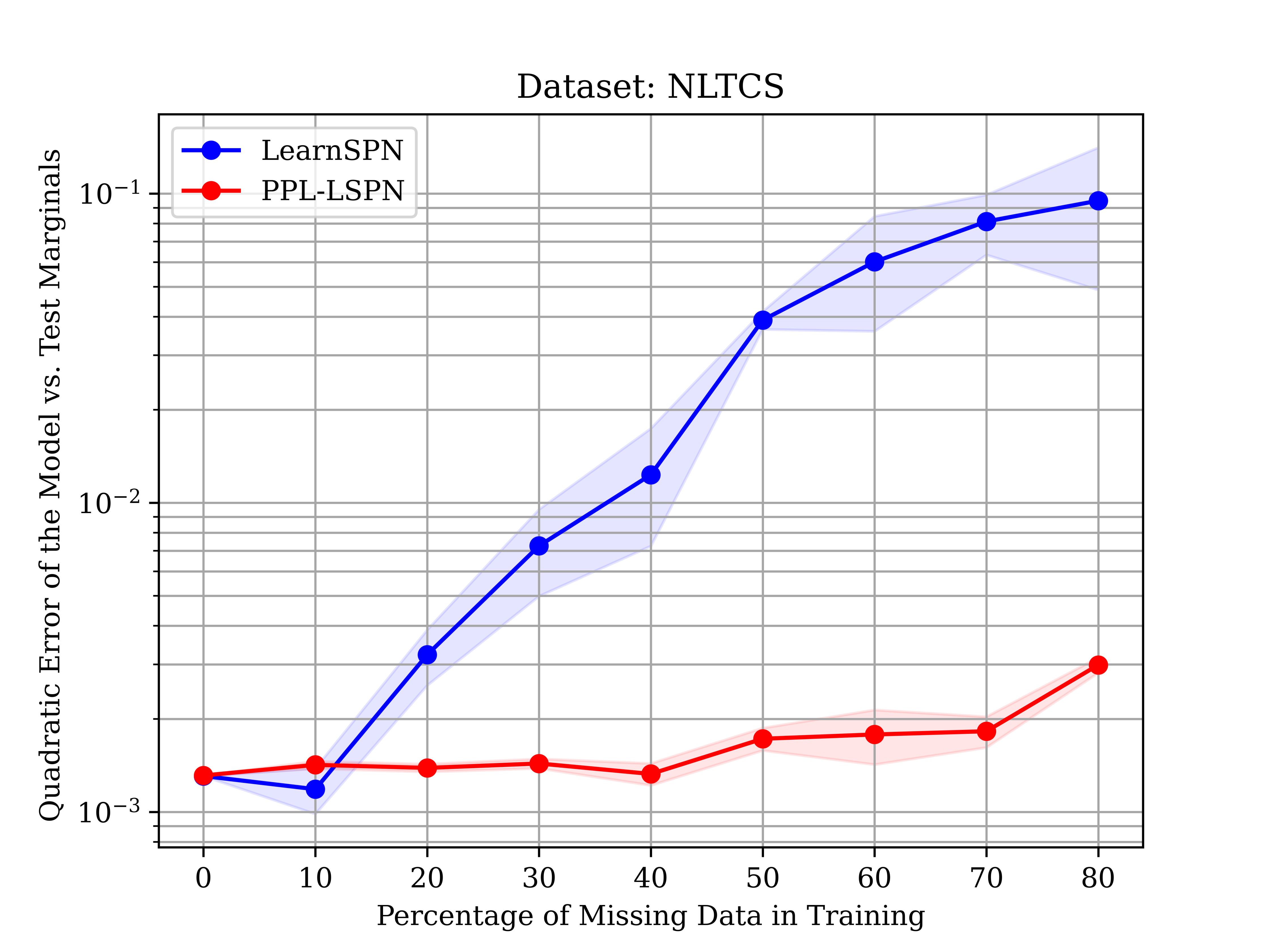}
    \end{subfigure}
     \begin{subfigure}[b]{0.32\textwidth}
         \centering
         \includegraphics[width=\linewidth]{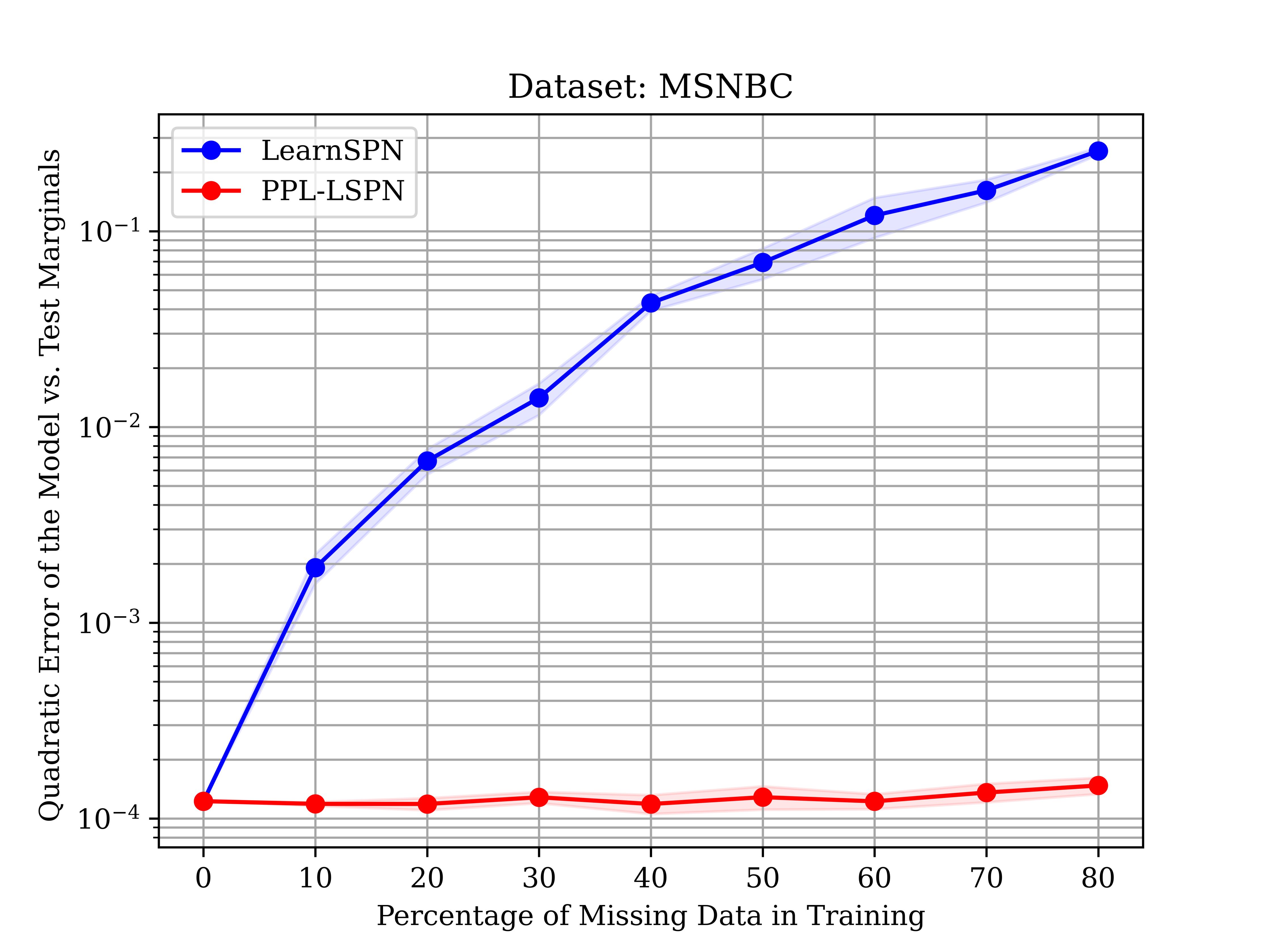}
     \end{subfigure}
     \begin{subfigure}[b]{0.32\textwidth}
         \centering
         \includegraphics[width=\linewidth]{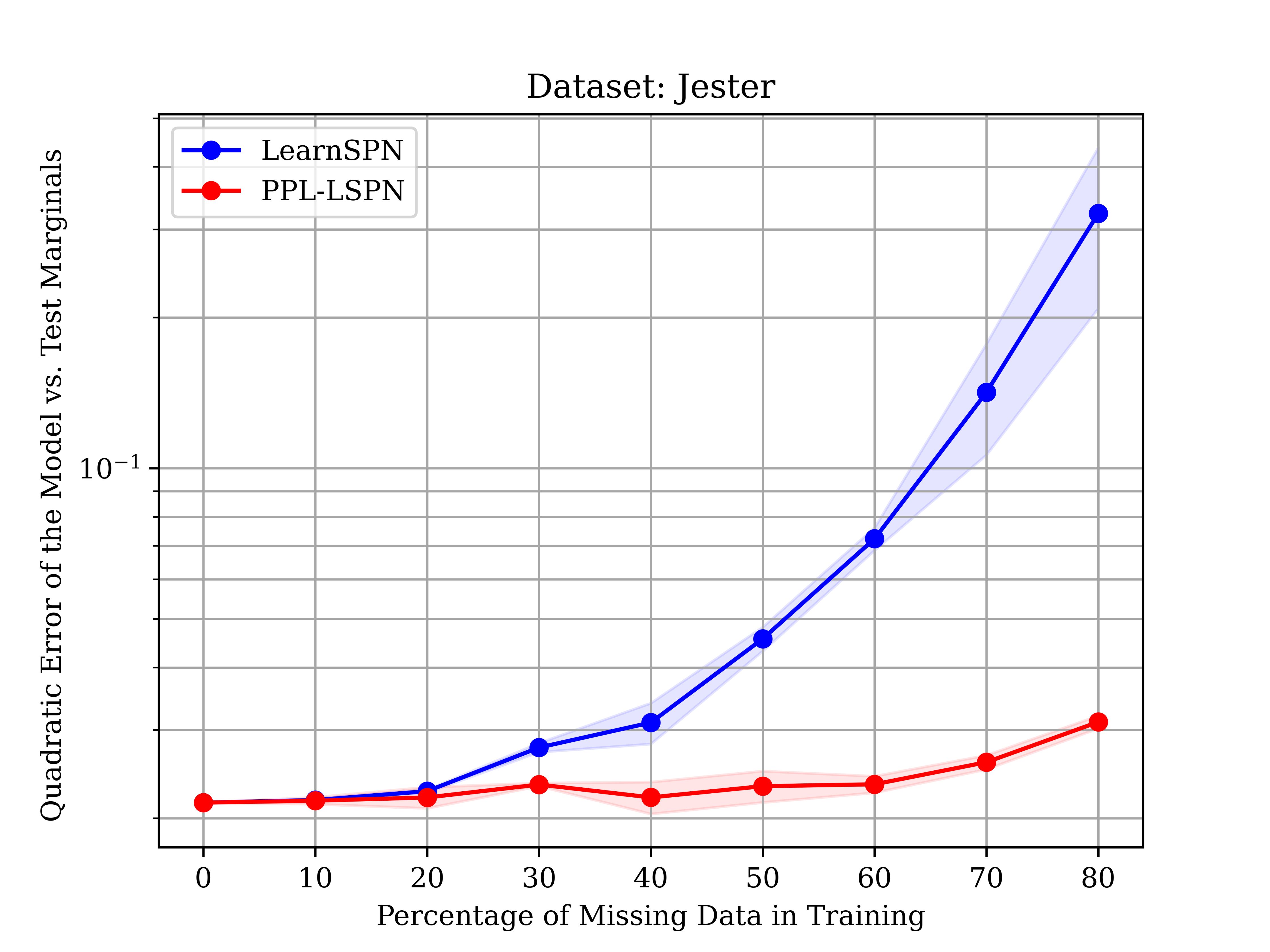}
     \end{subfigure}
    \caption{LearnSPN vs. (constrained) PPL-LSPN trained on datasets with MCAR missing values. Test log-likelihood measures the joint fitness, while quadratic error shows the quality of the marginals of the model with respect to the marginals of test data, with clear superior accuracy after constraints are imposed.}
     \label{fig:mv1}
 \end{figure*}

\subsection{Scarce datasets}\label{exp_scrc}

We carry out this experiment on three different binary datasets, namely NLTCS, MSNBC, and Jester \cite{lowd2010learning}. Our goal is to illustrate how additional information pertaining to the empirical marginal distributions can be incorporated into the PC so as to compensate for data scarcity. 
In order to simulate scarce data, we randomly subsample each dataset with a varying number of data instances. We use this subsample to train the PC model using LearnSPN and RAT-SPN. We then improve the model using the procedure described above so as to match the empirical marginal distributions: we add PPL constraints of the form $p(X_i=1)=\alpha_i$ for every variable $X_i$, which we enforce globally into the model. The test log-likelihood results are given in Figures \ref{fig:scrc-lspn} and \ref{fig:scrc_rspn}. The enhanced models obtained by applying the constrained optimization are called PPL-LSPN (variant of the LearnSPN baseline), and PPL-RSPN (variant of RAT-SPN). 

As can be seen, PPL-LSPN (Figure \ref{fig:scrc-lspn}) and PPL-RSPN (Figure \ref{fig:scrc_rspn}) slightly outperform LearnSPN (resp. RAT-SPN) as the training data become scarcer. This performance gain (in terms of testing data log-likelihood) is not similar across all datasets, which we attribute to the relative amount of information captured by the marginals. Arguably, in smaller datasets (in terms of the number of samples), matching marginals should lead to larger performance gains, as marginals encode a relatively larger amount of information. More importantly, matching marginals did not harm joint accuracy. 

We argue that marginal matching is even more advantageous to PPL-RSPN compared to PPL-LSPN, as the learned marginals are far more erroneous in the case of RAT-SPN. 
Figure \ref{fig:quad_err} displays the increase in quadratic error induced by \emph{not} matching marginals, for both LearnSPN and RAT-SPN. Clearly, a large gap can be observed between LearnSPN and RAT-SPN on scarce data. Somewhat to our surprise, estimated marginals in RAT-SPN are far off (also when compared to LearnSPN), making that model useless for marginal inference unless the constraints are imposed. Hence, and in particular for RAT-SPN, matching marginals leads to a strong improvement on the marginals themselves while (only but still) slightly improving the test joint likelihood.

\subsection{Experiments with missing values}\label{exp_mv}

We again use the three binary datasets above (NLCTS, MSNBC, and Jester). In order to simulate missing data, we train the baseline PC (via LearnSPN) in a missing completely at random (MCAR) setting, by removing entries completely at random from the data tables. After the models are trained, we enhance the learned distribution to match the training data marginals using the proposed approach.
Note that the current implementation of RAT-SPN mimics the effect of missing data with dropout layers (which is different from learning in presence of missing values); as well, the original version of RAT-SPN is not equipped to deal with missing data at training time, but can be easily tweaked for that purpose. We therefore focus these experiments on models trained with LearnSPN. 

Results with MCAR data are summarized in Figure \ref{fig:mv1}. The top plots show the joint testing data log-likelihood, while the bottom plots show the difference in the testing data marginal distributions (whose gains are very clear). We can see that in every experiment, as the proportion of missing values increases, the PC enhanced using constraints outperforms the base model, which suggests that marginal matching can be considered as an effective way to deal with missing data, potentially as an alternative to data imputation. 

\subsection{Fairness experiments}\label{exp_fair}

We investigate the impact of imposing fairness constraints in PCs. For each experiment, we assume variables $\X$ which comprise a binary class/target variable $Y \in \X$ and a binary protected attribute $X' \in \X, X'\neq Y$. Our objective is to improve the distribution learned via a PC towards fairness for the protected attribute when predicting class labels. We consider statistical parity as our measure of fairness (we use this as an example; we will not debate on fairness measures, since it is not the main focus of the paper). The corresponding fairness constraint is $p(y = 1 | x' = 1) = p(y = 1 | x' = 0)$. It is clear that this constraint is not of the form $\sum_{i}\tau_{i}\cdot p(F_{i}) \leq \alpha$. It would actually induce a non-linear constraint and the convex optimization could not be directly applied. However, we can lift the optimization problem to a higher dimension by including a new unknown $\beta$ where we take $p(y = 1 | x' = 1) = p(y = 1 | x' = 0) = \beta$. This latter can be decomposed into two separate linear constraints in the desired form:
\begin{equation}\label{decomp_const}
    \begin{split}
        &p((y = 1) \land (x' = 1)) - \beta\cdot p(x' = 1) = 0 , \\
        &p((y = 1) \land (x' = 0)) - \beta\cdot p(x' = 0) = 0 ; 
    \end{split}
\end{equation}
\noindent and as long as $\beta$ is fixed, the optimization can be carried out to impose the constraints in Equation~\eqref{decomp_const} to a learned PC using the proposed approach. In order to solve for $\beta$, we simply carry out an exhaustive search over candidate values between 0 and 1, retaining the best based on the performance of each resulting PC (obviously, this search procedure is reasonable for a single unknown $\beta$, or at most a few; otherwise, a smarter strategy would be required).

We consider six different classification datasets commonly used in fairness-aware machine learning, namely Adult, German Credit, Bank Marketing, Dutch Census, Credit Card Clients, and Law School \cite{le2022survey}. As in Section \ref{exp_scrc}, we refer to the variants as PPL-LSPN (for LearnSPN) and PPL-RSPN (for RAT-SPN). The details regarding the pre-processing of each dataset are provided in the appendix. The results are displayed in Tables \ref{tab:class_lspn} and \ref{tab:class_rspn}. Not only PPL-LSPN and PPL-RSPN are able to achieve a ``fair'' distribution w.r.t the protected attribute, but they also manage to do so without losing much of their representation power compared to LearnSPN or RAT-SPN, that is, the test likelihood and 0-1 accuracy are barely affected while statistical parity is enforced by the use of constraints. We stress out that our procedure being a post-processing of the PC at hand, models already trained and potentially in use in applications could be enhanced without the need of re-training from scratch. 

\begin{table}[h!]
    \centering
    \caption{Classification with LearnSPN vs. PPL-LSPN enforcing statistical parity via constraints.}
    \vskip 0.1in
    \begin{tabular}{|l|l|c|c|c|HHHHHcHH|}
        \toprule
         \multirow{2}{*}{\textbf{Dataset}}& \textbf{Protected} & \multirow{2}{*}{\textbf{Method}} & \multirow{2}{*}{\textbf{Test LL}} & \multirow{2}{*}{\textbf{Accuracy}} & \textbf{Balaced} & \textbf{TPR} & \textbf{TNR} & \textbf{TPR} & \textbf{TNR} & \textbf{Statistical} & \textbf{SP} & \textbf{Equalized} \\
         & \textbf{Attribute} &  &  &  & \textbf{Accuracy} & \textbf{prot.} & \textbf{prot.} & \textbf{non-prot.} & \textbf{non-prot} & \textbf{Parity} & \textbf{test} & \textbf{Odds} \\
         \hline
        \multirow{2}{*}{\textbf{Adult}} & \multirow{2}{*}{\textbf{Sex}} & LearnSPN & -13.614 & 0.8256 & 0.7269 & 0.3790 & 0.9746 & 0.5574 & 0.8909 & 0.1754 & 0.1846 & 0.2621 \\
         &  & PPL-LSPN & -13.764 & 0.7946 & 0.7144 & 0.4832 & 0.9558 & 0.5052 & 0.9085 & 0.0 & 0.1287 & 0.0736 \\
        \hline
        \textbf{German} & \multirow{2}{*}{\textbf{Sex}} & LearnSPN & -22.802 & 0.6993 & 0.5476 & 0.1313 & 0.9111 & 0.2027 & 0.9200 & -0.0171 & 0.0113 & 0.1283 \\
         \textbf{Credit} &  & PPL-LSPN & -23.075 & 0.704 & 0.5417 & 0.1615 & 0.9346 & 0.1840 & 0.8991 & 0.0 & 0.0257 & 0.1176 \\
         \hline
        \textbf{Bank} & \textbf{Marital} & LearnSPN & -16.448 & 0.8957 & 0.6112 & 0.2425 & 0.9784 & 0.2393 & 0.9835 & -0.0305 & -0.0128 & 0.0291 \\
         \textbf{Marketing} & \textbf{Status} & PPL-LSPN & -16.493 & 0.8949 & 0.5882 & 0.1796 & 0.9840 & 0.2054 & 0.9831 & 0.0 & -0.0034 & 0.0301 \\
         \hline
        \textbf{Dutch} & \multirow{2}{*}{\textbf{Sex}} & LearnSPN & -9.801 & 0.8141 & 0.8127 & 0.7284 & 0.9193 & 0.8095 & 0.7075 & 0.2520 & 0.3244 & 0.2929 \\
         \textbf{Census} &  & PPL-LSPN & -9.947 & 0.7359 & 0.7975 & 0.7588 & 0.8837 & 0.7124 & 0.8351 & 0.0 & 0.1776 & 0.0950 \\
         \hline
        \textbf{Cr. Card} & \multirow{2}{*}{\textbf{Sex}} & LearnSPN & -22.505 & 0.8164 & 0.6460 & 0.9388 & 0.3525 & 0.9549 & 0.3361 & 0.0185 & 0.0268 & 0.0394 \\
         \textbf{Clients} &  & PPL-LSPN & -22.539 & 0.8035 & 0.6443 & 0.9461 & 0.3408 & 0.9513 & 0.3383 & 0.0 & 0.0141 & 0.0266 \\
         \hline
        \textbf{Law} & \multirow{2}{*}{\textbf{Race}} & LearnSPN & -11.800 & 0.9076 & 0.5577 & 0.1679 & 0.9861 & 0.0819 & 0.9950 & -0.3012 & -0.0191 & 0.0950 \\
         \textbf{School} &  & PPL-LSPN & -11.845 & 0.9013 & 0.5606 & 0.1410 & 0.9901 & 0.1188 & 0.9931 & 0.0 & -0.0069 & 0.0433 \\
        \bottomrule
    \end{tabular}
    \label{tab:class_lspn}
    \vskip -0.1in
\end{table}

\begin{table}[htp!]
    \centering
    \caption{Classification with RAT-SPN vs. PPL-RSPN enforcing statistical parity via constraints.}
    \vskip 0.1in
    \begin{tabular}{|l|l|c|c|c|HHHHHcHH|}
        \toprule
         \multirow{2}{*}{\textbf{Dataset}}& \textbf{Protected} & \multirow{2}{*}{\textbf{Method}} & \multirow{2}{*}{\textbf{Test LL}} & \multirow{2}{*}{\textbf{Accuracy}} & \textbf{Balaced} & \textbf{TPR} & \textbf{TNR} & \textbf{TPR} & \textbf{TNR} & \textbf{Statistical} & \textbf{SP} & \textbf{Equalized} \\
         & \textbf{Attribute} &  &  &  & \textbf{Accuracy} & \textbf{prot.} & \textbf{prot.} & \textbf{non-prot.} & \textbf{non-prot} & \textbf{Parity} & \textbf{test} & \textbf{Odds} \\
         \hline
        \multirow{2}{*}{\textbf{Adult}} & \multirow{2}{*}{\textbf{Sex}} & RAT-SPN & -7.767 & 0.8193 & 0.7079 & 0.1610 & 0.9940 & 0.5437 & 0.8950 & 0.2000 & 0.2160 & 0.4817 \\
         &  & PPL-RSPN & -7.796 & 0.8148 & 0.7121 & 0.5056 & 0.9518 &  0.4989 & 0.9080 & 0.0 & 0.1136 & 0.0505 \\
        \hline
        \textbf{German} & \multirow{2}{*}{\textbf{Sex}} & RAT-SPN & -28.752 & 0.745 &  0.6419 & 0.28 & 0.8936 & 0.4848 & 0.8842 & -0.0309 & 0.0442 & 0.2142 \\
         \textbf{Credit} &  & PPL-RSPN & -28.756 & 0.745 & 0.6470 & 0.28 & 0.8936 & 0.5151 & 0.8736 & 0.0 & 0.0598 & 0.2550 \\
         \hline
        \textbf{Bank} & \textbf{Marital} & RAT-SPN & -13.736 & 0.8820 & 0.5539 & 0.2318 & 0.9612 & 0.0362 & 0.9963 & -0.0388 & -0.0578 & 0.2307 \\
         \textbf{Marketing} & \textbf{Status} & PPL-RSPN & -13.739 & 0.8788 &  0.5512 & 0.0227 & 0.9983 & 0.2051 & 0.9688 & 0.0 & 0.0450 & 0.2119 \\
         \hline
        \textbf{Dutch} & \multirow{2}{*}{\textbf{Sex}} & RAT-SPN & -12.880 & 0.7888 & 0.7885 & 0.7184 & 0.8997 & 0.8179 & 0.6066 & 0.2620 & 0.3546 & 0.3925 \\
         \textbf{Census} &  & PPL-RSPN & -12.923 & 0.7629 & 0.7583 & 0.7438 & 0.8535 & 0.6164 & 0.8622 & 0.0 & 0.0933 & 0.1361 \\
         \hline
        \textbf{Cr. Card} & \multirow{2}{*}{\textbf{Sex}} & RAT-SPN & -3.998 & 0.7838 & 0.5 & 1.0 & 0.0 & 1.0 & 0.0 & 0.0053 & 0.0 & 0.0 \\
         \textbf{Clients} &  & PPL-RSPN & -3.998 &  0.7838 & 0.5 & 1.0 & 0.0 & 1.0 & 0.0 & 0.0 & 0.0 & 0.0 \\
         \hline
        \textbf{Law} & \multirow{2}{*}{\textbf{Race}} & RAT-SPN & -7.274 & 0.9050 & 0.5154 & 0.0294 & 0.9960 & 0.0423 & 0.9936 & -0.2054 & 0.0030 & 0.0153 \\
         \textbf{School} &  & PPL-RSPN & -7.294 & 0.9034 & 0.5177 & 0.0294 & 0.9960 & 0.0529 & 0.9920 & 0.0 & 0.0055 & 0.0274 \\
        \bottomrule
    \end{tabular}
    \label{tab:class_rspn}
    \vskip -0.1in
\end{table}

\section{Conclusions and future work}\label{sec:concl}

We introduce a novel approach that allows to incorporate probabilistic propositional logic (PPL) constraints into a (pre-trained) probabilistic circuit (PC), so that the distribution encoded by the PC respects the constraints. We explain our design choices which allow for achieving tractable learning and inferences while ensuring that PPL constraints are satisfied. We also develop theoretical foundations that explain the feasibility of the optimization and how to reach an optimal solution in computationally tractable (polynomial) time. 
Experiments illustrate how we can take PCs and enhance them into better PCs that can be applied to practical scenarios, for example by applying fairness measures to the learned distribution and by (arguably) better handling missing values in the training data.

We make space for a couple of reflections. The goal of this research is to enhance machine learning models with probabilistic logic assessments, in the same spirit as neurosymbolic AI. 
We found out that PC models are already over-parameterized: thus, one can better tune the parameters in order to satisfy external constraints.
The first obvious idea is to do so via some variation of Expectation-Maximization or gradient methods, putting violation of constraints as (strong) penalties. However, it is not guaranteed that constraints are fully enforced; we therefore see that avenue as a great direction to investigate, even though the solution is likely to differ from the one described here. 
We managed to find a way to improve PCs a posteriori (without retraining) and efficiently (the optimization can run exactly and fast with modern convex optimization solvers). This choice comes at the expense of being able to only change the parameters of leaf distribution nodes; this---quite surprisingly---turns out to be enough to precisely enforce the constraints globally on the joint distribution while not losing model fitness. Moreover, we have no intention to claim that we are (or not) obtaining state-of-the-art results. This is an investigation of the combination of constraints into circuits, which we consider overall successful (but obviously not without limitations).
We see many possibilities with that. We are aware that the bucket size limitation is a serious complication, but creative experiments show that there may be many interesting problems to solve even under such limitation. Moreover, we know that the limitation can be mitigated by using some smarter parametrization of the local distributions: this direction is definitely worthwhile, although it may lead to a decrease in accuracy and will likely not provide the same guarantees as we currently have.

Beyond these research directions, the paper opens doors for future work, as the desire to combine probabilistic logic constraints and deep machine learning methods is immense. Possible immediate avenues include extending the applicability of constraints on continuous and mixed variables, applying constraints to new tasks such as other forms of fairness measures (for instance, equalized odds \cite{hardt2016equality}) in order to improve already learned PCs, improving the trade-off between accuracy and efficiency by using different optimizers, and considering extensions beyond consistent and valid PCs, to name but a few. 

\begin{credits}

\subsubsection{\discintname}
The authors have no competing interests to declare that are relevant to the content of this article.
\end{credits}

%
%
%
\bibliographystyle{splncs04}
\bibliography{refs}
%





\newpage
\appendix
\section{Data preprocessing for fairness experiments}

In this section, we explain the preprocessing measures that are applied to each dataset for fairness experiments. We would like to note that the specific structure of RAT-SPN requires a different set of preprocessing measures compared to LearnSPN. Being able to work with tensors restricts the type of data RAT-SPN can work with, mainly since tensors put specific restrictions on the parameter space of leaf nodes. As such, in addition to the general preprocessing measures (which are applied in both cases), datasets are also discretized and binarized (using one-hot encoding) for the case of RAT-SPN. 

\subsection{Adult dataset}

For the case where the base learner is LearnSPN, the instances containing missing values (3620 in total, equal to 7.41 \% of records) are removed from the dataset. The attribute \textit{fnlwgt}(final weight) is discarded from the dataset. Race is encoded as a binary attribute \textit{race} = \{\textit{white}, \textit{non-white}\}. The attribute \textit{age} is also discretized as \textit{age} = \{25-60, $<$25 or $>$60\}.

As for the case of RAT-SPN, in addition to the previous measures, the numerical attributes (\textit{capital gain}, \textit{capital loss}, \textit{hours per week}) are discretized as follows: \textit{capital gain} = \{$\leq$ 5000, $>$5000\}, \textit{capital loss} = \{$\leq$40, $>$40\}, \textit{hours per week} = \{$<$40, 40-60, $>$60\}. Additionally, categorical attributes are transformed as follows: \textit{workclass} = \{\textit{private}, \textit{non-private}\}, \textit{education} = \{\textit{high}, \textit{low}\}, \textit{marital-status} = \{\textit{married}, \textit{other}\}, \textit{relationship} = \{\textit{married}, \textit{other}\}, \textit{native-country} = \{\textit{US}, \textit{non-US}\}. Finally, the resulting dataset is discretized to be compatible with RAT-SPN.

\subsection{German dataset}

For the case where the base learner is LearnSPN, we extract gender information from attribute \textit{personal-status-and-sex} (which contains information on marital status and the gender of people), which leads to an additional attribute \textit{sex} (the protected attribute for our fairness experiments).

As for the case of RAT-SPN, additional transformations are as follows: \textit{duration} = \{$\leq$6, 7-12, $>$12\} (short, medium, and long-term); \textit{credit-amount} = \{$\leq$2000, 2000-5000, $>$5000\} (low, medium, and high income); \textit{age} = \{$\leq$25, $>$25\}. Finally, the resulting dataset is discretized to be compatible with RAT-SPN.

\subsection{Bank marketing dataset}

For the case where the base learner is LearnSPN, the only preprocessing measure is to extract a binary representation of marital status from attribute \textit{marital} as \textit{marital-bin} = \{\textit{married}, \textit{non-married}\}. the attribute \textit{marital-bin} is added to the original dataset as an additional attribute, representing the protected group for our fairness experiment.

As for the case of RAT-SPN, additional transformations are as follows: \textit{job} = \{\textit{blue-collar}, \textit{management}, \textit{service}, \textit{other}\}; \textit{balance} = \{$\leq$0, $>$0\}; \textit{day} = \{$\leq$15, $>$15\}; \textit{duration} = \{$\leq$120, 121-600, $>$600\}; \textit{campaign} = \{$\leq$1, 2-5, $>$5\}; \textit{pdays} = \{$\leq$30, 31-180, $>$180\}; \textit{previous} = \{0, 1-5, $>$5\}; \textit{age} = \{25-60, $<$25 or $>$60\}. Finally, the resulting dataset is discretized to be compatible with RAT-SPN.

\subsection{Dutch census dataset}

For this dataset, no particular preprocessing has been done. For LearnSPN, the dataset is utilized in its original form, and for the case of RAT-SPN, the only process is to binarize the dataset to address compatibility issues.

\subsection{Credit card clients dataset}

For the case where the base learner is LearnSPN, the only preprocessing is to drop the attribute \textit{id}, as it does not contain any useful information about the task. For the of RAT-SPN, additional transformations are as follows: \textit{age} = \{$\leq$35, 36-60, $>$60\};  the amount of the given credit (\textit{limit\_bal}),the amount of the bill statements (\textit{bill\_amt\_1},...,\textit{bill\_amt\_6}), and the amount of the previous payments (\textit{pay\_amt\_1},...,\textit{pay\_amt\_6}) = \{$\leq$50000, 50001-200000, $>$200000\} (corresponding to \textit{low}, \textit{medium}, \textit{high} levels); history of the past payments \textit{pay\_0}, ..., \textit{pay\_6} = \{\textit{pay dully}, \textit{1-3 months}, $>$\textit{3 months}\}.

\subsection{Law school dataset}

For the case where the base learner is LearnSPN, the dataset is utilized in its original form. For the case of RAT-SPN, additional transformations are as follows: \textit{decile1b} = \{$\leq$5, $>$5\}, \textit{decile3} = \{$\leq$5, $>$5\}, \textit{lsat} = \{$\leq$37, $>$37\}, \textit{ugpa} = \{$<$3.3, $\geq$3.3\}, \textit{zgpa} = \{$\leq$0, $>$0\}, \textit{zfygpa} = \{$\leq$0, $>$0\}.

\end{document}